\documentclass{article}
\pdfoutput=1
\usepackage[nonatbib]{neurips_2022}
\usepackage{microtype}
\usepackage{graphicx}
\usepackage{subfigure}
\usepackage{booktabs} 
\usepackage{subfigure}
\usepackage{algorithm}
\usepackage{enumitem}
\usepackage{algorithmic}

\usepackage{tabu}
\usepackage{multirow}
\usepackage{multicol}
\usepackage{wrapfig}
\usepackage{amsmath}
\usepackage{amssymb}
\usepackage{mathtools}
\usepackage{amsthm}
\usepackage{lipsum}
\usepackage{float}

\PassOptionsToPackage{numbers, compress}{natbib}

\DeclareMathOperator*{\argmin}{arg\,min}
\DeclareMathOperator*{\argmax}{arg\,max}

\newcommand{\set}[1]{\mathcal{#1}}

\newcommand{\std}{\scriptsize$\pm$}

\newcommand{\tblue}[1]{{\color{blue}#1}}

\newcommand{\inda}{\text{InDA}}
\newcommand{\exda}{\text{ExDA}}
\newcommand{\da}{\text{DA}}

\newcommand{\old}{\text{old}}

\usepackage[utf8]{inputenc} 
\usepackage[T1]{fontenc}    
\usepackage{hyperref}       
\usepackage{url}            
\usepackage{booktabs}       
\usepackage{amsfonts}       
\usepackage{nicefrac}       
\usepackage{microtype}      
\usepackage{xcolor}         

\title{Efficient Scheduling of Data Augmentation \\ for Deep Reinforcement Learning}

\author{%
  Byungchan Ko\thanks{This work was done while Byungchan Ko studied in GSAI, POSTECH.  }
  \\NALBI
  \\kbc@nalbi.ai
  \And
  Jungseul Ok 
  \\GSAI, POSTECH
  \\jungseul@postech.ac.kr
}

\begin{document}

\maketitle
\begin{abstract}

In {deep reinforcement} learning (RL), data augmentation is widely considered as a tool to induce a set of useful priors about semantic consistency and to improve sample efficiency and generalization performance. However, even when the prior is useful for generalization, distilling it to RL agent often interferes with RL training and degenerates sample efficiency. Meanwhile, the agent is forgetful of the prior due to the non-stationary nature of RL. These observations suggest two extreme schedules of distillation: (i) over the entire training; or (ii) only at the end. Hence, we devise a stand-alone network distillation method to inject the consistency prior at any time (even after RL), and a simple yet efficient framework to automatically schedule the distillation. Specifically, the proposed framework first focuses on mastering train environments regardless of generalization by adaptively deciding which {\it or no} augmentation to be used for the training. After this, we add the distillation to extract the remaining benefits for generalization from all the augmentations, which requires no additional new samples. In our experiments, we demonstrate the utility of the proposed framework, in particular, that considers postponing the augmentation to the end of RL training.
\href{https://github.com/kbc-6723/es-da}{\tblue{https://github.com/kbc-6723/es-da}}

\end{abstract}

\section{Introduction}


Deep reinforcement learning (RL) aims at finding 
a neural network to represent policy or value functions
taking raw observation as input, of which the most common form in practice is visual data or images of high-dimensionality, e.g., video games \cite{mnih2015human}, 
board games \cite{silver2017mastering, silver2018general}, and robot controls \cite{tobin2017domain, kalashnikov2018qt}. 
RL handling high-dimensional input often suffers from 
poor sample efficiency and generalization capability, 
mainly due to the curse of dimensionality \cite{bellman2015adaptive, henderson2019deep}.
To overcome these issues,
it has been widely considered to augment data
based on prior knowledge that a set of transformations preserve 
the meaning or context of input observations, e.g.,
cropping out unimportant parts of images, and changing colors
\cite{laskin2020reinforcement, kostrikov2020image, lee2019network, hansen2021generalization}.
On one hand,
RL agent 
can be directly fed with the original and augmented data
so that it implicitly learns a representation with the prior
and improves the sample efficiency and generalization \cite{laskin2020reinforcement}.
On the other hand, 
the prior knowledge in data augmentation
can be explicitly distilled via a self-supervised learning,
which introduces additional regularization
to ensure consistency between responses to original and augmented inputs \cite{raileanu2020automatic, hansen2021generalization}.



However,
data augmentation shows highly task-dependent effect in RL,
and it is prone to generate severe interference with the training
even when it truly conveys a useful prior to train and test environments
\cite{laskin2020reinforcement, raileanu2020automatic, hansen2021generalization}.
We address the problem of alleviating the interference between data augmentation and RL training to improve {\it sample efficiency} for acquainting train tasks, and {\it generalization capability} for unseen test environments.
This problem in (online) RL is more critical and challenging than
that in supervised learning (SL)
since the objective function and
data distribution are time-varying in RL,
while they are not in SL.
Indeed, according to 
\cite{achille2018critical,golatkar2019time},
it is sufficient to partly apply data augmentation 
just for a short period of SL at any time.
Meanwhile, we empirically observe that the prior from data augmentation
can be easily forgotten in RL of the non-stationary nature
(see Section~\ref{sec:scheduling}), i.e.,
the effect of augmentation is {\it time-sensitive}.

Based on our observation about the interference and the time-sensitivity, 
we propose two simple yet effective methods according to timings of data augmentation :
{\bf In}tra {\bf D}istillation with {\bf A}ugmented observations (\inda{})
and {\bf Ex}tra {\bf D}istillation with {\bf A}ugmented observations (\exda{}).
Data augmentation beneficial for the sample efficiency 
needs to be applied over the entire RL training, i.e., \inda{}.
Conversely, data augmentation useful only for the generalization
should be postponed to the end of RL training, i.e., \exda{},
so that we can minimize the interference in the training,
while enjoying the benefit in the testing. 
\inda{} and \exda{} are equipped with 
{\bf D}istillation with {\bf A}ugmented observation (DA).
\da{} is a stand-alone self-supervised learning 
which enables us to induce the prior after RL training,
and shows a relatively small interference with RL training
by explicitly preserving the response of RL agent to the original input. 

The best timing (\inda{} or \exda{}) depends on traits of train task and augmentation.
We hence suggest a framework of adaptive scheduling, named UCB-\exda{},
that 
(i) first aims at
maximizing the sample efficiency by adaptively 
selecting which or no augmentation to be used {\it during RL training};
and then, (ii) 
distill the priors from all the augmentations {\it after RL training}.
Specifically, 
inspired by \cite{raileanu2020automatic},
we devise UCB-\inda{}
for the adaptive selection in the first part by modifying the upper confidence bound (UCB) algorithm \cite{auer2002using} for multi-armed bandit,
where differently from \cite{raileanu2020automatic}, 
the set of arms includes all the augmentations and {\it the option to not augment}.
In summary, UCB-\exda{} is nothing but executing UCB-\inda{} followed by \exda{}.
Our experiment demonstrates the utility of the proposed framework.



Our contributions are summarized as follows:
\begin{enumerate}
\label{contribution}
\item[(i)] 
We devise \da{} (Section~\ref{sec:distillation}) which explicitly preserves the knowledge of RL agent
so that enables distilling the consistency prior of augmentation into RL agent not only during but also after RL training,
while other methods 
\cite{laskin2020reinforcement, raileanu2020automatic, hansen2021generalization}
need to be applied concurrently with RL training
and thus show relatively strong interference in our experiments (Section~\ref{sec:DA_benefit}).


\item[(ii)] 
We identify the simple yet effective timings of data augmentation: either \inda{} or \exda{} (Section~\ref{sec:inda}, Section~\ref{sec:exda}), based upon
the discovery of the time-sensitivity (Section~\ref{sec:scheduling})
that has not been observed in SL \cite{golatkar2019time}, i.e., the proposed timings are effective particularly for RL.





\item[(iii)] We finally establish UCB-\exda{} which automatically decides the best timing of augmentation, and demonstrate its superiority compared to others (Section~\ref{sec:eval}).
The advantage of UCB-\exda{} is particularly substantial when the best strategy is \exda{} postponing augmentation to the end of RL training.

\end{enumerate} 

\section{Related Works}

\smallskip 
\noindent
{\bf Augmented experience in RL.}
To solve the problem of
poor generalization and sparse data,
a popular approach is
to fabricate virtual experiences 
and train the RL agent to learn with them.
Domain randomization
is a technique to produce such experiences 
from a simulator of a targeted system,  \cite{tobin2017domain, pinto2017asymmetric, raparthy2020generating}.
Accurate simulators of practical systems are difficult to obtain, and it has limited applicability.
However, visual augmentation has no such limit because the method uses simple image transformations
such as cropping, tilting, and color jitter,
although applications require a careful understanding of the targeted system 
to guide the design of an appropriate image transformer.
A method of curriculum learning for domain randomization, in which the difficulty is gradually increasing \cite{raparthy2020generating}
provided insights that coincide with some of our findings.
However, we provide a further understanding of the types of visual augmentation that should be early or late during training.

Regularization from augmented data in vision-based RL
has been implemented in various learning frameworks,
including but not limited to 
representation \cite{hansen2021generalization, stooke2020decoupling},
self-supervised \cite{raileanu2020automatic},
and contrast \cite{srinivas2020curl}. 
One proposed algorithm \cite{raileanu2020automatic} applies the UCB algorithm \cite{auer2002using} to automatically select the most effective  augmentations over RL training, where each augmentation is considered as an arm and then evaluate the effectiveness of augmentation by using a sliding window average.
The idea of adapting augmentation concurs with our main message
regarding the timing of augmentation.
In \cite{raileanu2020automatic}, 'not augmenting' is not an option, whereas our findings indicate that it should be. In addition, \cite{raileanu2020automatic} does not consider post augmentation followed by RL training, as in \exda{}.

\smallskip 
\noindent
{\bf Other time-sensitivity in deep learning.} 
During deep learning, the early stage of training often has a significant effect \cite{erhan2010does,achille2018critical}.
Therefore, we devised time-sensitive
methods that adapt to the progress of training, such as learning rate decay \cite{you2019does} and curriculum learning
\cite{wu2021curricula}. 
Golatkar {\it et al}. \cite{golatkar2019time} studied such a time-sensitivity of 
regularization techniques for SL, where 
the effect of data augmentation in different time
does not change much. 
We find that 
the time-sensitivity of augmentation can be significant in RL.
This contrast may occur because of the non-stationary nature of RL, 
which SL does not have.
Although a set of techniques originally developed for SL such as convolutional neural network, weight decay, batch normalization, dropout and self-supervised learning
improve deep RL 
\cite{higgins2018darla,cobbe2019quantifying,liu2020regularization, farebrother2020generalization, srinivas2020curl,yarats2020improving, hansen2020self},
a thorough study should be conducted before introducing 
a method from different learning frameworks, because we find the contrasting time-sensitivities of data augmentation.
This spirit is also shared with 
an application \cite{igl2020impact} of implicit bias in SL
\cite{gunasekar2017implicit,arora2019implicit,gidel2019implicit} to RL.

\section{Background} \label{sec:back}

\paragraph{Notation.} We consider a standard agent-environment interface
of vision-based reinforcement learning in a discrete 
Markov decision process of state space $\set{S}$,
action space $\set{A}$, and
kernel $P = P(s_{t+1}, r_{t}  | s_t, a_t)$
which determines the state transition and reward distribution.
The goal of the RL agent is to find a policy 
that maximizes the expectation of cumulative reward $\sum_{t=0}^{t'-1} \gamma^{t} r_{t}$, where $t'$ is terminating time and 
$\gamma \in [0, 1]$ is discount factor.
At each timestep $t$, the agent selects an action $a_t \in \set{A}$
and receives a reward $r_t$
and an image 
$o_{t+1} = O(s_{t+1}) \in \mathbb{R}^{k \times k'}$
as an observation of the next state $s_{t+1}$.
Data augmentation can be described by an image transformation function 
$\phi: \mathbb{R}^{k \times k'} \mapsto \mathbb{R}^{k \times k'}$
of which output is presumed to have the same or similar semantics of the input.

\smallskip
\paragraph{Baseline RL algorithm.} 
Throughout this paper, we use 
Proximal Policy Optimization (PPO) \cite{schulman2015high} as a baseline, 
although we believe our findings and methods
can be easily adjusted to others.
PPO is a representative on-policy RL algorithm to learn policy $\pi_\theta (a \mid o)$ and value function $V_\theta$ of neural agent parameterized by $\theta$.
Storing a set of recent transitions $\tau_t := (o_t, a_t, r_t, o_{t+1})$ in experience buffer $\set{D}$, the agent 
is updated to minimize the following loss function:
\begin{align} \label{eq:ppo}
   L_{\text{PPO}}(\theta) 
   = - L_{\pi}(\theta) + \alpha L_{V}(\theta) \;,
\end{align}
where $\alpha$ is a hyperparameter and 
canonical regularization terms are omitted.
The clipped policy objective function
$L_{\pi}$ and value loss function $L_{V}$
are defined as:
\begin{align}
\!\!\!\!\!L_{\pi}(\theta)  
&= \hat{\mathbb{E}} \! 
\left[\text{min}(\rho_t(\theta)\hat{A}_t, \text{clip}(\rho_t(\theta), 1-\epsilon, 1+\epsilon)\hat{A}_t) \right] \!\!\! 
\\
\!\!\!\!\!L_{V}(\theta)  
&= \hat{\mathbb{E}}\! 
\left[
\left(V_\theta(o_t) - V_t^{\text{targ}}\right)^2\right] \;,
\end{align}
where the expectation $\hat{\mathbb{E}}$
is
taken with respect to $\tau_t \sim \set{D}$, 
we denote by
$\theta_{\text{old}}$ the parameter before the update,
$\rho_t(\theta)$
is the importance ratio $\frac{\pi_{\theta}(a_t|o_t)}{\pi_{\theta_{\text{old}}}(a_t|o_t)}$,
and $\hat{A}_{t}$ is the generalized advantage estimation \cite{schulman2015high}.



\section{Method}
\label{sec:method}

\begin{figure}[!t]
    \begin{minipage}{0.47\textwidth}
    \centering
    \centering
        \includegraphics[width=\textwidth]{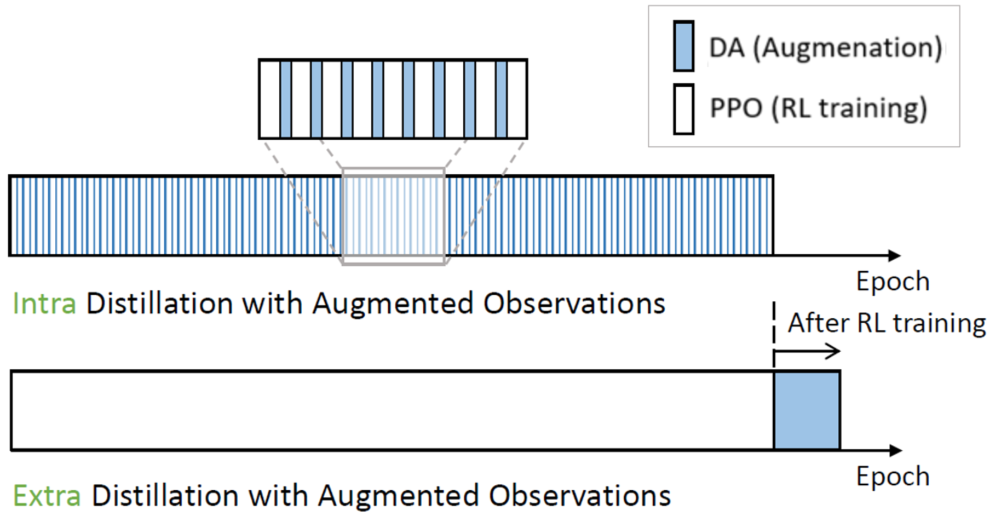}
        \caption{An illustration of \inda{} and \exda{} 
        \label{fig:inda-vs-exda}}
    \end{minipage}
    \hspace{0.3cm}
    \begin{minipage}{0.47\textwidth}
    \centering
        \includegraphics[width=0.75\textwidth]{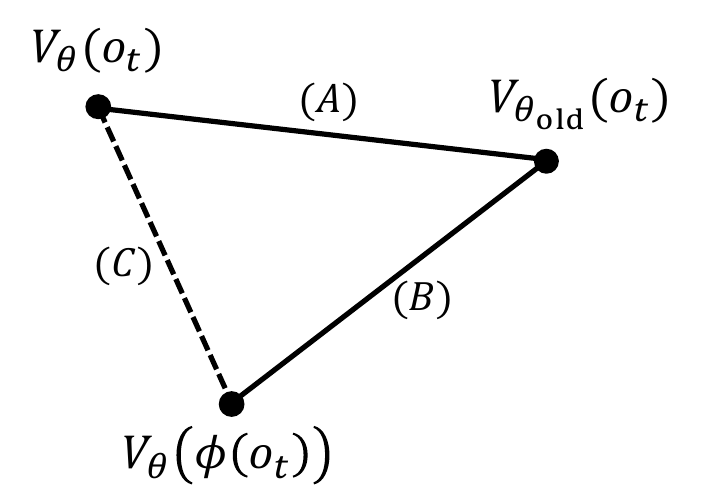}
        \caption{An illustration of distillation losses.
        \label{fig:discrepancy}}
    \end{minipage}
\end{figure}

\smallskip
\noindent 
In what follows, we present our methods: 
{\bf D}istillation with {\bf A}ugmented observations (\da{}; Section~\ref{sec:distillation}),
{\bf In}tra {\bf DA} (\inda{}; Section~\ref{sec:inda}), 
{\bf Ex}tra {\bf DA} (\exda{}; Section~\ref{sec:exda}), and 
then 
the adaptive scheduling frameworks based on UCB (UCB-\inda{} and UCB-\exda{}; Section~\ref{sec:ucb}).
\da{} 
is a stand-alone knowledge distillation method, which can be used at any time to instill the underlying prior of augmentation into a given RL agent.
\inda{} and \exda{} conduct
either \da{} or PPO in each epoch but have different schedules (Figure~\ref{fig:inda-vs-exda}),
where
\inda{} 
interleaves PPO and \da{},
whereas \exda{} 
performs PPO first then \da{}.
UCB-\inda{} 
adaptively decides
which or no augmentation 
to be used in each \da{} epoch of \inda{}
based on the UCB of estimated gain from each option.
UCB-\exda{} performs \exda{} preceded by UCB-\inda{}.




\subsection{Distillation with augmented observations (\da{})}
\label{sec:distillation}

The underlying prior of augmentation can be infused by minimizing a measure of inconsistency between the agent's responses to original and augmented inputs (resp. $o_t$ and $\phi(o_t)$). 
For instance, with PPO agent learning policy $\pi_\theta$ and value $V_\theta$,
Raileanu {\it et al}. \cite{raileanu2020automatic} proposes 
the following measure:
\begin{align} \label{eq:drac}
L_{\text{dis}}(\theta, \phi) 
:=
\hat{\mathbb{E}}_{o_t \sim \set{D}} 
\left[ \text{KL}[\pi_{\theta}(\cdot |o_t), \pi_\theta(\cdot |\phi(o_t))] \right]
+ \hat{\mathbb{E}}_{o_t \sim \set{D}}
\left[(V_{\theta}(o_t)-V_{\theta}(\phi(o_t)))^2 \right] \;,
\end{align}
which uses Kullback-Leibler divergence (first term) and mean squared deviation (second term) for policy and value inconsistencies, respectively.
Noting that $L_{\text{dis}}(\theta; \phi)$ can be minimized to be zero by a constant response to all inputs,
the distillation with Eq~\eqref{eq:drac} can distort the RL agent, in particular, when applying it outside of RL training. 


We hence devise a network distillation technique (\da{})
which {\it explicitly preserves} the RL agent's response to original input and thus is applicable even after RL training.
\da{} distills the knowledge of RL agent $\theta_{\text{old}}$ into $\theta$ using not only original but also augmented observations.
More formally, the loss of \da{} is written as:
\begin{align} \label{eq:L-DA}
L_{\da{}}(\theta, \phi; \theta_{\text{old}}) := 
L_{\text{dis}}(\theta, \mathbb{I}; \theta_{\text{old}}) +
L_{\text{dis}}(\theta, \phi; \theta_{\text{old}}) \;.
\end{align} 
We here denote the identity transformation by $\mathbb{I}$
such that $\mathbb{I}(o) = o$ for all $o$,
and extend the definition of $L_{\text{dis}}$ in Eq~\eqref{eq:drac} as follows:
\begin{align}
\label{eq:L-dist}
L_{\text{dis}}(\theta, \phi; \theta_{\text{old}}) 
&= \hat{\mathbb{E}}_{o_t \sim \set{D}} \left[ \text{KL}[\pi_{\theta_{\text{old}}}(\cdot |o_t), \pi_\theta(\cdot |\phi(o_t))] \right]
+ \hat{\mathbb{E}}_{o_t \sim \set{D}}
\left[(V_{\theta_{\text{old}}}(o_t)-V_{\theta}(\phi(o_t)))^2 \right]     \;.
\end{align}

In Eq~\eqref{eq:L-DA}, the first term ensures 
that $\theta$ and $\theta_{\text{old}}$ behave identically
for the original inputs, and the second one infuses the consistency prior.
In Figure~\ref{fig:discrepancy},
$L_{\text{dis}}(\theta, \mathbb{I}; \theta_{\text{old}})$,
$L_{\text{dis}}(\theta, \phi; \theta_{\text{old}})$,
and $L_{\text{dis}}(\theta, \phi; \theta)$
graphically correspond to (A), (B), and (C), respectively.
From this, it follows that 
minimizing $L_{\text{\da{}}}$ in Eq~\eqref{eq:L-DA} 
eventually reduces $L_{\text{dis}}(\theta, \phi; \theta)$ in Eq~\eqref{eq:drac}. 
In addition, 
the minimization of $L_{\da{}}$
secures
the responses of $\theta$ to 
the original inputs (which can be pre-computed to reduce computation cost)
to the those of $\theta_{\text{old}}$, 
while the alternatives (e.g., (A)+(C):
$L_{\text{dis}}(\theta, \mathbb{I}; \theta_{\text{old}}) +
L_{\text{dis}}(\theta, \phi; \theta)$)
does not and thus may generate interference with RL training. 
Our experiments (Section~\ref{sec:DA_benefit}; Figure~\ref{fig:robust}) justifies the design of \da{}
by showing substantial advantage compared to
the other alternatives such as DrAC \cite{raileanu2020automatic} using $L_{\text{dis}}(\theta, \phi; \theta)$ in Eq~\eqref{eq:drac}. We note that this advantage becomes much clearer when a wrong augmentation is given, c.f., the supplementary material.





\subsection{Intra distillation with augmented observations} 
\label{sec:inda}

\inda{} (Algorithm~\ref{alg:inda}) alternates between minimizing 
$L_{\text{PPO}}$ and $L_{\da{}}$, i.e., 
PPO and DA are explicitly separated,
whereas they are often executed simultaneously in other methods 
\cite{raileanu2020automatic}.
Such a clear separation reduces the interference \cite{hansen2021generalization}.
We can control the frequency and timing of applying distillation with hyperparameters 
$I$,
$S$ and $T$, 
where we perform \da{} after each $I$ rounds of RL training
only if the current epoch $n$ is in the interval of $[S,T]$, i.e., 
$S$ and $T$ are the epochs to begin and terminate \da{}, respectively.
We vary $S$ and $T$ to study the time-dependency
of data augmentation.
We denote \inda{}$[S, T]$ to indicate the period to apply \da{},
while we omit the indication when \da{} is applied over the entire period.
We provide further details on \inda{} in the supplementary material.

\begin{minipage}{0.47\textwidth}
\begin{algorithm}[H]
\caption{\inda{} \label{alg:inda}}
\begin{algorithmic}[1]
\REQUIRE  
            $N, I, \phi$, $S$, $T$ 
            \smallskip
            \STATE  Initialize $\theta$ close to origin.
            \FOR{$n =  1, 2, \ldots, N$}
                \STATE // RL training
                    \STATE Store sampled transitions to $\set{D}$;
                    \STATE Optimize RL objective ${L}_{\text{PPO}}(\theta)$
                    with $\set{D}$;
                \smallskip
                \STATE // Distillation 
                \IF{ $n \in [S, T]$ and $\text{mod}(n-1, I) = 0$}  
                    \STATE Store $\theta_{\text{old}} \leftarrow \theta$;
                    \STATE Minimize $L_{\da{}}(\theta)$
                    for $\set{D}$, $\theta_{\text{old}}$ and $\phi$;
                \ENDIF
            \ENDFOR 
\end{algorithmic}
\end{algorithm}
\end{minipage}
\hfill
\begin{minipage}{0.47\textwidth}
\begin{algorithm}[H]
\caption{\exda{} \label{alg:exda}}
\begin{algorithmic}[1]
\REQUIRE $N$, $M$, $\phi$
            \smallskip
            
            \STATE  Initialize ${\theta}$ close to origin. 
            \STATE //Pre-training phase with RL algorithm    	
            \FOR{$n = 1, 2, \ldots, N$ }
                \STATE Store sampled transitions to $\set{D}$;
                \STATE Optimize RL objective ${L}_{\text{PPO}}(\theta)$
                with $\set{D}$;
            \ENDFOR 
            \STATE
            \smallskip
            \STATE Store $\theta_{\text{old}} \leftarrow \theta$;
            \STATE // Distillation at the end of RL training
            \FOR{$m = 1, 2, \ldots, M$ }
                \STATE Minimize $L_{\da{}}(\theta)$ for 
                $\set{D}$, $\theta_{\old{}}$ and $\phi$;
            \ENDFOR 
\end{algorithmic}
\end{algorithm}
\end{minipage}

\subsection{Extra distillation with augmented observations} 
\label{sec:exda}

\exda{} (Algorithm~\ref{alg:exda}) 
performs the distillation after the end of RL training, 
where the lengths of \da{} and RL training are parameterized by 
$M$ and $N$, respectively.
We note that 
computational cost can be reduced by removing the value inconsistency measure $\hat{\mathbb{E}}_{o_t \sim \set{D}}
\left[(V_{\theta_{\text{old}}}(o_t)-V_{\theta}(\phi(o_t)))^2 \right]$
from $L_{\text{dis}}$ in Eq~\eqref{eq:L-dist}
because 
 the value consistency is not necessary for RL training nor distillation in the actor-critic framework and including it has a potential risk of generating additional interference.
In the supplementary material,
it is empirically verified that this reduction does not degrade RL performance.
We leave more interesting details 
in the supplementary.
For instance, 
one can consider re-initializing 
$\theta$ before starting \da{}
as a part of exploiting the implicit bias \cite{gunasekar2017implicit, igl2020impact} 
to improve generalization.
However, test performance is often dropped.
This is mainly because 
the dataset to distill $\pi_{\theta_{old}}$ has much less diversity than that observed along the trajectory.
Thus, we use no re-initialization for the experiments in the main paper. 


\subsection{
UCB-based adaptive scheduling frameworks
} \label{sec:ucb}

The training benefit by augmentation differs depending on the task. This dependency complicates the choice of whether to use \inda{} or \exda{} for augmentation. 
Hence, we devise an auto-augmentation method, called UCB-\inda{},
inspired by UCB-DrAC \cite{raileanu2020automatic},
where each augmentation is corresponded
to an arm in
multi-armed bandit (MAB) problem
and assessed its potential gain in training with
upper confidence bound (UCB) \cite{auer2002using}.
More formally, 
in UCB-\inda,
the set of arms is the set of augmentations, $\Phi = \{ \phi_1, \dots , \phi_K \}$, which must include the identity function $\mathbb{I}$, i.e., the option not to augment. The inclusion of identity function is small but makes crucial
difference than UCB-DrAC \cite{raileanu2020automatic}
since we observe that using augmentation sometimes needs to be postponed
after RL training for the sake of better sampling complexity and test performance.

A round of MAB corresponds to every $I$ epoch of \inda{}, where we let $\phi_{k(s)}  \in \Phi$ be the augmentation selected at the $s$-th round of \da{}. Let $G(s)$ be the average return, 
the sum of estimated advantage $\hat{A}$ and predicted value $V_\theta$, over 
$(I-1)$ epochs of PPO followed by the $s$-th \da{}.
UCB algorithm assumes that each arm generates reward independently drawn from a fixed distribution,
and estimates the empirical mean over the entire sampling process.
However, in RL, the return $G(s)$ is non-stationary,
so UCB-\inda{} computes moving average gain $\bar{G}_{k}(s)$, instead,
taken over a certain number (chosen to be 3 in our experiment) of most recent rounds selecting $\phi_k$ as Raileanu {\it et al}. \cite{raileanu2020automatic} proposed. 
Then, inspired by UCB1 algorithm \cite{auer2002using}, UCB-\inda{} 
selects action $k(s)$ at round $s$ as follows:
\begin{equation}
    k(s) = \argmax_{k \in \{1, ..., K\}} \bar{G}_{k} (s) + c \sqrt{\frac{\log(s)}{N_k (s)}}
\end{equation}
where $c$ is the UCB exploration coefficient, and $N_k(s)$ is the number of
times selecting $\phi_k$ up to round $s$.
We refer to the supplementary material
for the hyperparameter choice.
We remark that compared to UCB-DrAC \cite{raileanu2020automatic},
the proposed UCB-\inda{} 
has subtle but important differences, summarized in two folds:
(i) the inclusion of identity transformation (i.e., no augmentation)
and (ii) the distillation with augmentation via \inda{}.
The gain of each component is numerically studied in Section~\ref{sec:exp}.
Finally, we note that UCB-\exda{} is nothing but UCB-\inda{} followed by \exda{}.


\subsection{PAGrad}
\label{sec:pagrad}

\begin{figure*}[t!]
    \centering
    \centering
    \subfigure[non-conflicting: $\langle g_{\text{aux}}, g_{\text{main}} \rangle \ge 0$]{
        \includegraphics
        [ height=0.12\textheight]{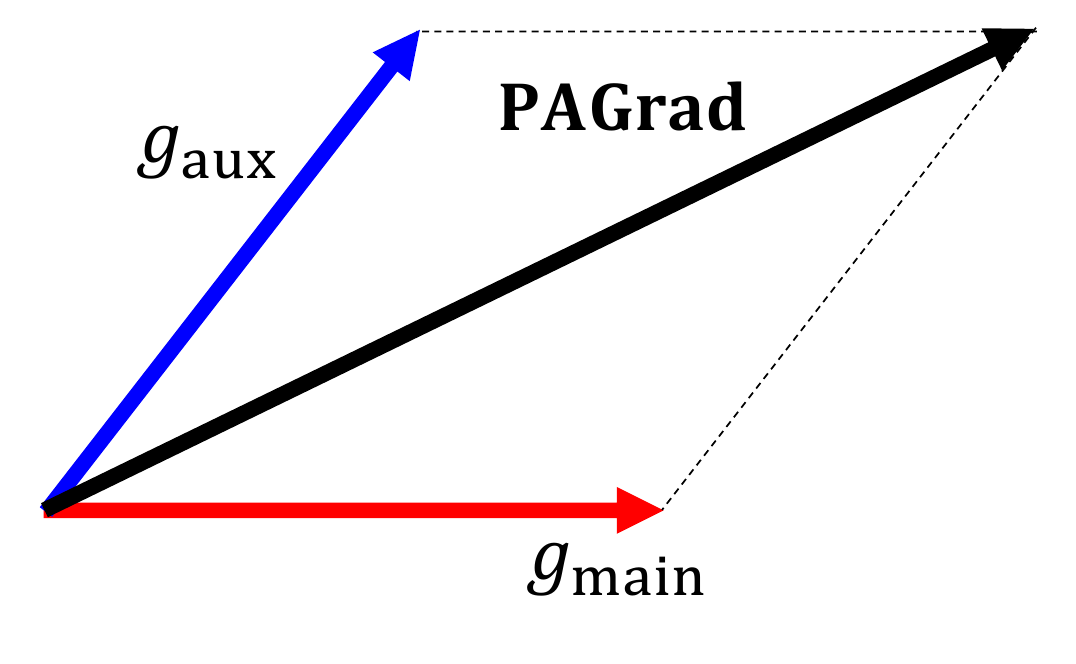}
        \label{fig:pj1}
        }
        \hspace{0.5cm}
    \subfigure[conflicting: $\langle g_{\text{aux}}, g_{\text{main}} \rangle < 0$ ]{
        \includegraphics[height=0.12\textheight]{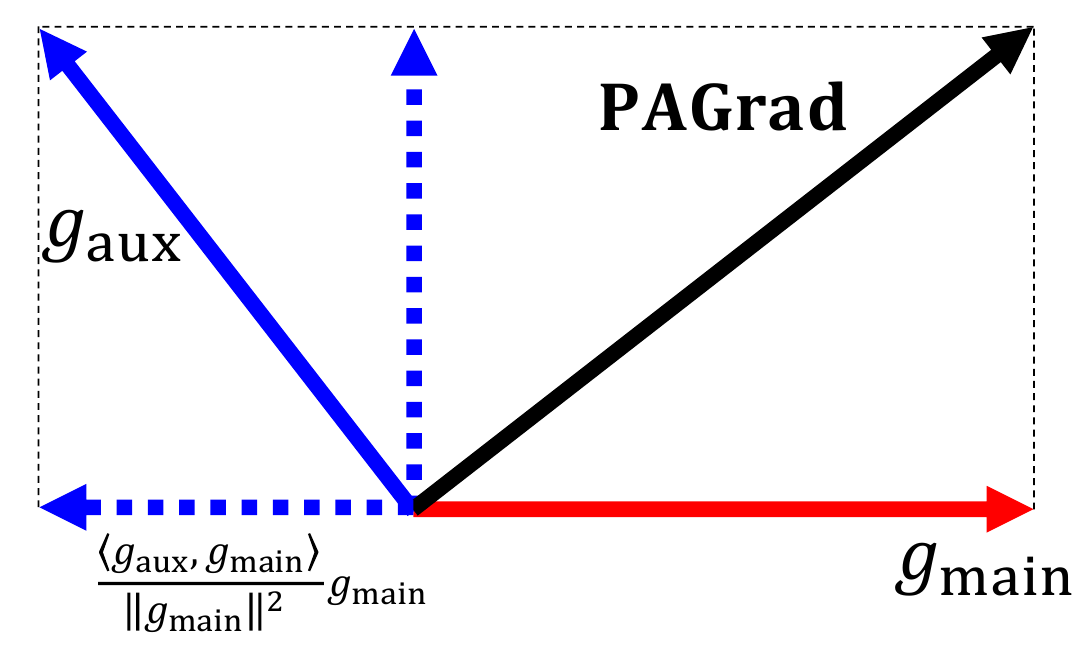}
        \label{fig:pj2}
        }
    \caption{
    {\it An illustration of gradient conflicting and PAGrad.}
    We here let $g_{\text{main}}$
    and $g_{\text{aux}}$ denote the gradients of 
    main (shown in red; e.g., $\nabla L_{\text{PPO}}$) and auxiliary
    (shown in blue; e.g., $\nabla L_{\text{dis}}$) losses, respectively.
    In the computation of
    PAGrad (shown in black), 
    the component
    of $g_{\text{aux}}$ 
    conflicting to $g_{\text{main}}$ 
    only when it exists (i.e., $\langle  g_{\text{aux}}, g_{\text{main}} \rangle < 0$).
    }
    \label{fig:projection}
\end{figure*}

Inspired by
\cite{yu2020gradient},
we devise an alternative approach to reduce the interference
by interpreting 
RL training with data augmentation
as a multi-task learning,
where 
$L_{\text{PPO}}$
and $L_{\text{dis}}$
correspond
to the main and auxiliary
task losses, respectively.
In \cite{yu2020gradient},
the degree of conflict from task A
to task B is estimated 
by 
the inner product of
the gradient of task A
and 
the negative gradient of task B, c.f.,
Figure~\ref{fig:projection}.
From this, we propose
PAGrad (Projecting Auxiliary Gradient)
to compute
a modified gradient 
of $L_{\text{PPO}}$
excluding
the conflict
from the auxiliary
gradient
$\nabla L_{\text{dis}}$
to 
the main 
one
$\nabla L_{\text{PPO}}$.
Formally, PAGrad computes
the gradient given as follows:
\begin{align} \label{eq:PAGrad}
\nabla L_{\text{PPO}} 
+\left(
\nabla L_{\text{dis}}
- 
\tfrac{
\min\{0, \langle \nabla L_{\text{dis}}, \nabla L_{\text{PPO}} \rangle \} }{\| \nabla L_{\text{PPO}}  \|^2}  \nabla L_{\text{PPO}} 
\right)  \;,
\end{align}
where 
$\tfrac{ \min\{0, \langle \nabla L_{\text{dis}} , \nabla L_{\text{PPO}} \rangle \} }{\| \nabla L_{\text{PPO}}  \|^2}  \nabla L_{\text{PPO}}$
is
the components of $\nabla L_{\text{dis}}$
opposite to $\nabla L_{\text{PPO}}$
which may disturb optimizing the main objective $L_{\text{PPO}}$.
Based on this,
we devise 
DrAC+PAGrad
that updates the model parameter toward the negative direction
of \eqref{eq:PAGrad}.
This is an alternative
of \inda{}, while
it concurrently optimizes
$L_{\text{PPO}}$ and $L_{\text{dis}}$
differently from \inda{}
adopting the time-separation
of optimizing
$L_{\text{PPO}}$ and $L_{\text{dis}}$.
We also note that 
it differs from
the original method proposed in \cite{yu2020gradient}
alternating
the main and auxiliary tasks to accomplish
every task at equal priority, while
we have a clear priority
on RL task.

\newcommand{\rowfont}[1]{
   \gdef\rowfonttype{#1}#1%
}

\section{Experiment}
\label{sec:exp}

\vspace{-0.5em}

\subsection{Setups}
\label{sec:setup}
\vspace{-0.5em}
\paragraph{Train and test tasks.}
We use the OpenAI Procgen benchmark 
of 16 video games
\cite{cobbe2020leveraging}, 
where a main character 
tries to achieve a specific goal, 
e.g., finding exit (Maze) 
or collecting coins (Coinrun), while avoiding enemies given a 2D map.
At each time $t$, visual observation $o_t$ is given as an image of size $64 \times 64$.
A train or test task 
is to achieve a high score on a set of environments
configured by game and mode,
where a mode describes predefined 
sets of levels (e.g., complexity of map) and backgrounds.
Cobbe {\it et al}. \cite{cobbe2020leveraging} provide {\it easy} mode for each game,
consisting of $200$ levels and a certain set of backgrounds.

We simplify {\it easy} mode 
and
train agents in {\it easybg} mode, of which the only difference from {\it easy} mode \cite{cobbe2020leveraging} is 
showing only a single background.
This is useful to investigate the case that using visual augmentation is helpful for testing but not for training.
We denote the task configuration by game\_name({\it mode}), e.g., Coinrun({\it easybg}).
Then, we evaluate generalization capabilities
using two modes:
{\it test-bg} and {\it test-lv},
which contain unseen backgrounds and levels, respectively, in addition to {\it easybg} mode that we use for training.


\vspace{-0.5em}
\paragraph{Types of augmentation.}
For clarity, we mainly focus on two types of visual augmentation,
each of which conveys distinguishing inductive bias:
\begin{enumerate}[label=(\alph*)]
\item {\it Random color} transforms an image by passing through either color jitter layer or random convolutional layer
proposed in \cite{lee2019network}. 
From this, we can impose the consistency to color changes, which may provide a strong generalization to diverse backgrounds of {\it test-bg} mode.

\item {\it Random crop} leaves a randomly-selected rectangle
and pads zeros to the rest of the image \cite{raileanu2020automatic}.
This augmentation is particularly useful in fully-observable games (e.g., Chaser and Heist), because it imposes an efficient attention mechanism.
\end{enumerate}
We also report the result with other augmentations
including {\it color jitter, random convolution, gray, and cutout color} in the supplementary material, from which the same messages can be interpreted.

\vspace{-0.5em}
\paragraph{Baseline methods for RL with data augmentation.}
We mainly compare the proposed methods (\inda{} and \exda{})
to the following baselines:
\begin{enumerate}[label=(\alph*)]
\item {\it RAD} \cite{laskin2020reinforcement} simply feeds
PPO with experiences of original and augmented observations. 
\item  {\it DrAC} \cite{raileanu2020automatic} 
is a method to simultaneously minimizing $L_{\text{PPO}}$ in Eq~\eqref{eq:ppo} and $L_{\text{dis}}$ in Eq~\eqref{eq:drac}.

\item {\it DrAC+PAGrad}
is a variant of DrAC, which we devise to
investigate another mechanism to relieve the interference between RL training and augmentation in Section~\ref{sec:pagrad}.

\end{enumerate}

The supplementary material presents further details
and experiments, 
which we omitted for simplicity.
All results in the main paper are averaged over five runs.

\begin{figure}[H]
    \centering
    \includegraphics[width=0.58\linewidth]{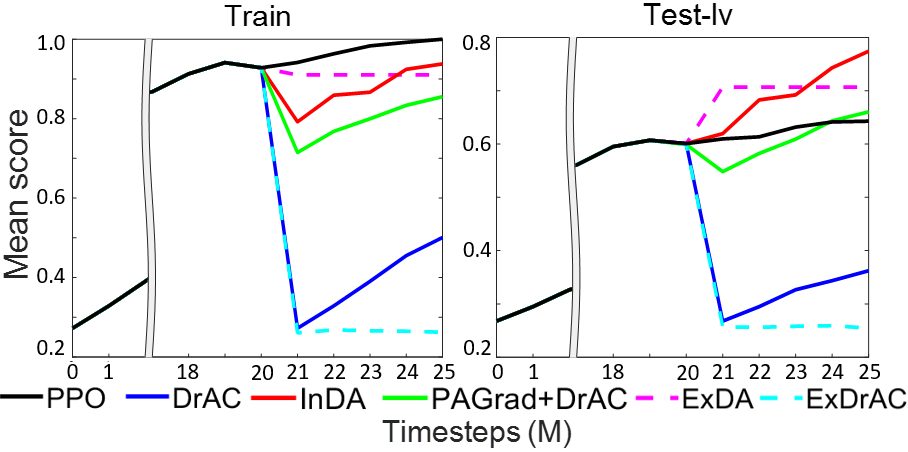}
    \caption{ {\it Benefit of separating distillation from RL training.} We compare \inda{}, \exda{}, DrAC, DrAC+PAGrad and ExDrAC 
    when we start to apply each of them
    to distill the prior of {random crop}
    after 20M timesteps of PPO. ExDrAC is a variant of DrAC
    without RL training, i.e., minimizing only
    $L_{\text{dis}}$ in Eq~\eqref{eq:drac}.
    }
    \label{fig:robust}
\end{figure}
\subsection{Benefit of separating distillation from RL training.}
\label{sec:DA_benefit}

In Figure~\ref{fig:robust}, 
after training PPO agent up to 20M timesteps, 
we start to suddenly apply one of the different distillation methods with {random crop}. 
We report averaged scores over 6 environments (Bigfish, Dodgeball, Plunder, Chaser, Heist, Maze) after normalized by the highest train score of PPO on each environment.
After 20M timesteps,
\exda{} and ExDrAC have no RL training 
but minimize
$L_{\da{}}$ in Eq~\eqref{eq:L-DA} and $L_{\text{dis}}$ in Eq~\eqref{eq:drac}, respectively. 
The substantial gap between \exda{} and ExDrAC
justifies the design of \da{} explicitly preserving the knowledge from RL when distilling the prior.
More importantly, 
it is remarkable that \exda{} promptly learns the generalization ability once it starts 
to distill the prior. This validates the idea of postponing
the distillation after RL training.

We now compare the distillation methods
(\inda{}, DrAC, and DrAC+PAGrad)
concurrently optimizing the RL objective 
and distilling the prior in Figure~\ref{fig:robust}.
Each method has performance degeneration due to the 
interference generated by distillation.
However, \inda{} and DrAC+PAGrad have clearly smaller degeneration than DrAC 
which is the only one without any separation of optimizing the RL objective and distillation loss.
We note that 
DrAC+PAGrad has more interference than 
\inda{}, and it seems to fail to impose the prior
since there is not much difference to PPO 
in testing. 
Hence, this verifies the superiority of \inda{}
which enables distilling the prior while alleviating the interference.

\begin{figure}[t]
    \centering
    \subfigure[train (easybg)]{
        \includegraphics[width=0.3\linewidth]{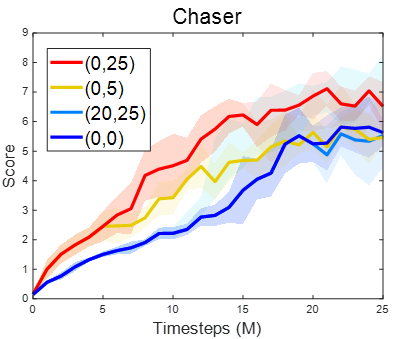}
        \label{fig:chaser_train}
        }   
    \subfigure[test-lv]{
        \includegraphics[width=0.3\linewidth]{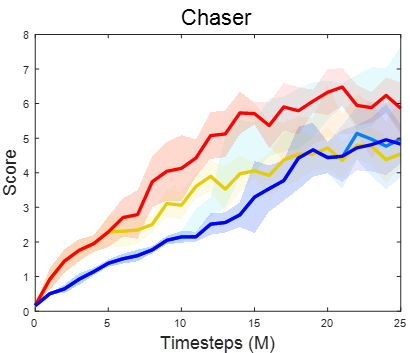}
        \label{fig:chaser_test}
        }
    \subfigure[Policy Distance]{
        \includegraphics[width=0.3\linewidth]{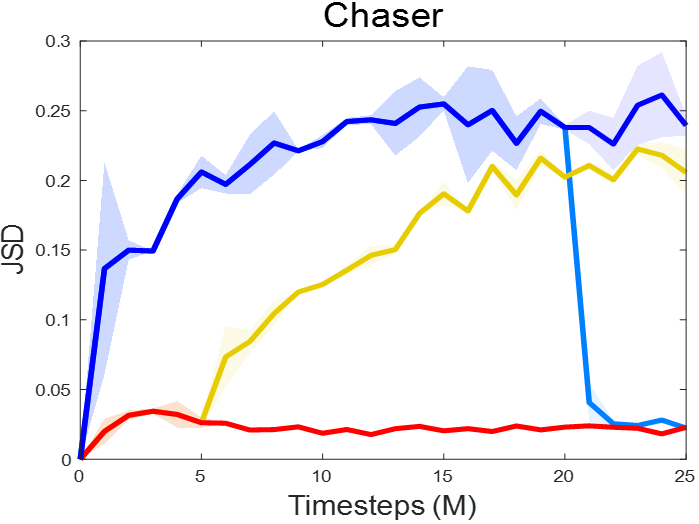}
        \label{fig:chaser_jsd}
        }
    
    \subfigure[train (easybg)]{
        \includegraphics[width=0.3\linewidth]{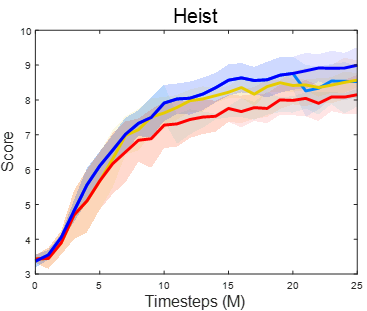}
        \label{fig:heist_train}
        }   
    \subfigure[test-lv]{
        \includegraphics[width=0.3\linewidth]{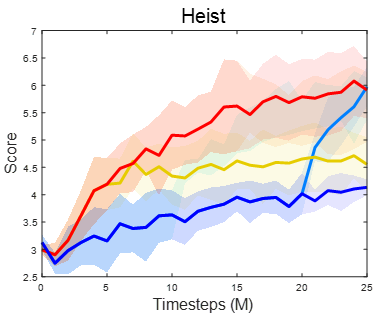}
        \label{fig:heist_test}
        }
    \subfigure[Policy Distance]{
        \includegraphics[width=0.3\linewidth]{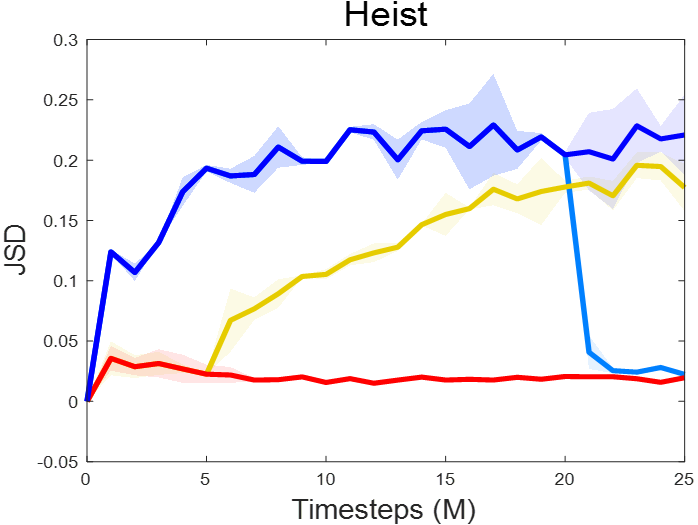}
        \label{fig:heist_jsd}
        }
    \caption{{\it Time-sensitivity of applying augmentation.} We compare train and test performance of \inda$[S, T]$ with {\it random crop},
    where the timing of applying \da{} is governed by starting time $S$ and terminating time  $T$, and we evaluate four different pairs of $[S, T] = [0, 0], [0, 5], [20, 25], [0, 25]$ up to 25M timesteps. Furthermore, we show the change of distance between two policies on an original observation and an augmented observation. 
    Note that \inda{} with $[S, T]= [0, 0]$ means RL training with no augmentation, i.e., vanilla PPO.
    We focus on Chaser and Heist
    since they exemplify two representative time-dependencies.
    Each train task uses {\it easybg} mode.
    We present further details and results with other games in the supplementary material.
    \label{fig:time_sensitivity}
    \vspace{-1em}
    }

\end{figure}

\subsection{Effective timings of distillation}
\label{sec:scheduling}



In what follows, we aim to identifying
effective timings of distillation. 
To this end, 
in Figure~\ref{fig:time_sensitivity},
we test several schedules of applying \da{}
on two representative environments of distinguishing traits.
The supplementary material
presents the result on more environments,
which is similar one of the representatives.

\paragraph{An effective timing: \inda{}.}

Chaser with {random crop} (Figure~\ref{fig:chaser_train}~and~\ref{fig:chaser_test})
represents the case when augmentation improves the sample efficiency of training
and thus the generalization ability in training. To compare the generalization ability, we measure policy distance between original and augmented observations using Jenson-Shannon divergence (Figure~\ref{fig:chaser_jsd}). \inda[0, 5]'s policy distance is increased after it stops using \da{}. Thus, the generalization ability is degraded, if we do not continue to use \da{}.
In this case, it is clear that \inda{} should be applied 
during the entire RL training, 
as \inda[0, 25] shows the best performance in both training and testing.
In addition, it is also important to apply
\da{}
as soon as possible
since
the effect of \inda[0,5] applying \da{} in the beginning
is relatively prompt and significant comparing
to that of \inda[20,25]. 
This suggests that the automatic framework needs to explore more in the early stage. 

\paragraph{An effective timing: \exda{}.}


Heist with {\it random crop} (Figure~\ref{fig:heist_train}~and~\ref{fig:heist_test})
shows the opposite use of data augmentation 
to what Chaser case suggests, i.e.,
postponing data augmentation as much as possible. 
Random crop generates a slight interference,
although it immediately improves the generalization ability.
We remark that
the inductive bias from the random crop is easily forgotten, 
as the test performance of \inda{}[0, 5] is saturated right after stopping the distillation. 
This can be explained by the time-varying nature of sample distribution 
and objective in RL.
Interestingly, it is in contrast to the data augmentation in SL, 
where an early distillation is sufficient to impose the prior \cite{golatkar2019time}.
On the other hand,
the test performance curve of \inda{}[20, 25]
soars right after \da{}.
Furthermore, Figure~\ref{fig:heist_jsd} shows that \inda{}[20, 25] narrows the gap between two policies on the original and augmented observation. 
Recalling the interference between RL training and distillation,
this suggests postponing the distillation at the end of RL training,
and motivates our \exda{}.

\paragraph{Performance benchmark on \inda{} and \exda{}.}
In Table~\ref{table:Performance Benchmark All}, 
we summarize the train and test performances 
of \inda{} and \exda{}
on a set of games with different augmentations and modes.
\exda{} outperforms other baselines with {\it random color} on {\it test-bg}. Moreover,
we note that \exda{} consumes only 0.5M timesteps to inject the prior at the end of RL training, whereas the others use all the training data. 
\exda{} in both sample efficiency and generalization ability with {\it random crop} on {\it test-lv}. It is worth noting that DrAC+PAGrad is slightly better than DrAC, while there is a substantial gap between \inda{} and
each DrAC-based algorithm. This again confirms the benefit from the separated distillation of \inda{} observed in Figure~\ref{fig:robust}.
These results demonstrate that each combination of environment and augmentation has a suitable time at which to apply augmentation, and 
the gain from using the right distillation timing, i.e., 
online (\inda{}, DrAC, or DrAC+PAGrad) or offline (\exda{}),
is rigid regardless of the choice of algorithms.

\begin{table*}[ht]
\tiny
    \centering
    \renewcommand{\multirowsetup}{\centering}
    \resizebox{0.92\textwidth}{!}{
      \begin{tabular}{ ccc|c|cc|ccc}
        \toprule  
        \multicolumn{2}{c}{Augmentation} & Task & PPO  & RAD  & DrAC  & DrAC+PAGrad & \inda{} &  \exda{}\\
        \midrule  
        \multirow{6}{5mm}{Rand color}& \multirow{3}{10mm}{Rand conv} & Train & \bf{1.00}	& 0.98 & 0.88 & 0.89  & 0.88 & 0.98 \\
        &                         & Test-bg  & 1.00   & 1.08 & 1.86 & 1.88 & 1.92 & \bf{2.11}  \\
        &                         & Test-lv & \bf{1.00}  & 0.81 & 0.84 & 0.84 & 0.7 & 0.87 \\
        \cmidrule(lr){2-9}
        & \multirow{3}{10mm}{Color jitter} & Train & \bf{1.00} & 0.94 & 0.95 & 0.95 & 0.96 & 0.98  \\
        &                         & Test-bg  & 1.00	  & 1.37 & 1.44 & 1.44 & 1.43 & \bf{1.48} \\
        &                         & Test-lv & \bf{1.00}   & 0.83  & 0.86 & 0.86 & 0.84 & 0.88 \\
        \midrule
        \multicolumn{2}{c}{ \multirow{3}{15mm}{Rand crop}}  & Train & 1.00  & 0.28 & 1.08& 1.09 & \bf{1.25} & 0.91  \\ 
        &                         & Test-bg & \bf{1.00} & 0.64 & 0.87  & 0.94 & 0.94 & 0.95 \\
        & & Test-lv  & 1.00	  & 0.46  & 1.52 & 1.53& \bf{1.80} & 1.09 \\
        \bottomrule
      \end{tabular}}
    \caption{{\it Benchmark of \inda{} and \exda{}.} 
    We report normalized  train and test scores of \inda{}, \exda{} and DrAC with PAGrad on Open AI Procgen, compared to baselines PPO, DrAC \cite{raileanu2020automatic}, RAD \cite{laskin2020reinforcement}. {\bf Boldface} indicates the best performance.
    We average the score among several environments [(Rand color: coinrun, ninja, climber, fruitbot, jumper, heist, maze), (Rand crop: Bigfish, Dodgeball, Plunder, Heist, Chaser, Maze)] after normalization considering PPO score to be $1$.
    Every method is trained on 200 levels, using {\it easybg} mode. We evaluate test performance on both {\it test-bg} and {\it test-lv}. The results can be interpreted
    as an upper bound of potential gain from using data augmentation at perfect timing.
    }
    \label{table:Performance Benchmark All} 
\end{table*}

\subsection{Adaptive scheduling methods}
\label{sec:eval}

\begin{figure}[b]
    \centering
    \subfigure[Chaser]{
        \includegraphics[width=0.33\linewidth]{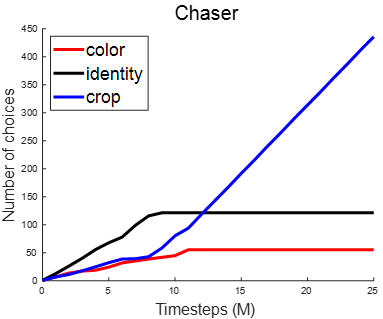}
        \label{fig:ucb_chaser}
        }
        \hspace{0.5cm}
    \subfigure[Heist]{
        \includegraphics[width=0.33\linewidth]{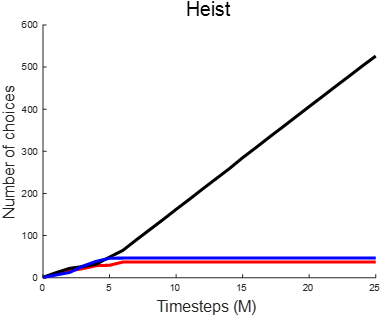}
        \label{fig:ucb_heist}
        }

    \caption{{\it Exploration \& exploitation to find the most beneficial augmentation}. We show that selected augmentation during training with UCB-InDA for each Chaser and Heist. UCB-\inda{} automatically selects the augmentation among three options, random color, identity and random crop.    }
    \label{fig:ucb_selection}
\end{figure}

\begin{table*}[!t]
\tiny
    \centering
    \renewcommand{\multirowsetup}{\centering}
    \resizebox{\textwidth}{!}{
      \begin{tabular}{ c|c|c|cc|cccc}
        \toprule
        \multirow{1}{5mm}{Env}& Method & PPO & DrAC & UCB-DrAC  & InDA  & ExDA  & UCB-InDA & UCB-ExDA  \\

        \midrule  
        \multirow{3}{5mm}{Heist} & Train   & 9.2  $\pm$ 0.46 & 3.53 $\pm$0.3 & 7.41$\pm$ 2.09  & 4.9 $\pm$ 0.79 & 9.14  $\pm$ 0.56 & \bf{9.67}$\pm$ 0.23 & 9.45     $\pm$ 0.29  \\
                                 & Test-bg & 5.18 $\pm$ 1.53&3.58 $\pm$ 0.31 & 3.76$\pm$ 0.54 & 4.9 $\pm$ 0.87& 7.05 $\pm$ 1.29 & 6.23$\pm$ 1.29 & \bf{7.86}$\pm$ 0.83 \\
                                 & Test-lv & 4.13 $\pm$ 1.39 & 3.07 $\pm$ 0.33 & 3.49$\pm$0.48  & 1.47$\pm$ 0.77 & 5.05 $\pm$ 0.98 & 4.8$\pm$       1.24& \bf{5.74}$\pm$ 0.67   \\
                                 
        \midrule  
        \multirow{3}{5mm}{Chaser} & Train & 5.63$\pm$ 1.12 & 0.16$\pm$0.02 & 4.6$\pm$ 1.22  & 5.69$\pm$ 1.42 & 5.58$\pm$ 1.33 &  \bf{7.49}$\pm$1.27 &7.23$\pm$ 1.15  \\
                                 &Test-bg  & 0.87$\pm$ 0.06 & 0.1$\pm$ 0.02& 0.57$\pm$0.12  & 3.51$\pm$ 1.33 & 1.02$\pm$ 0.04 & 1.08$\pm$ 0.08 & \bf{3. 18}$\pm$ 0.79 \\
                                 &Test-lv & 4.83$\pm$ 0.88 &0.14$\pm$0.01 & 4.14$\pm$1.01  & 4.96$\pm$ 1.21& 5.11$\pm$  1.05 &  \bf{6.45}$\pm$ 0.8 & 6.43$\pm$ 1.28  \\
        
        \bottomrule
      \end{tabular}}
    \caption{{\it Full exploitation of  augmentation to improve generalization on both test-bg and test-lv}. We compare \inda{}, \exda{}, DrAC, UCB-DrAC, UCB-InDA, UCB-ExDA and PPO about train, test-bg and test-lv. {\bf Boldface} indicates the best method. \inda{} and DrAC use both random color and random crop during RL training. \exda{} use both augmentation after RL training. UCB-\inda{} and UCB-DrAC are trained as automatically selecting the augmentation during training. UCB-\exda{} trains \exda{} after UCB-\inda{} with both augmentation. }
    \label{table:full_exploit} 
\end{table*}

\paragraph{Adaptive selecting of timings: UCB-\inda{} and UCB-\exda{}.}
It is hard to know in advance whether a certain augmentation helps the training or not \cite{raileanu2020automatic}.
We hence employ UCB-\inda{} which estimates
the gain or damage from each augmentation
from trial and error
and identifies the one with most help.
Recall
\autoref{table:full_exploit} 
where 
PPO (without augmentation) shows much better training performance than \inda{} in Heist.
As shown in Figure~\ref{fig:ucb_heist},
UCB-\inda{} is able to identify that no augmentation 
is best for training in Heist.
This implies that \exda{} is more appropriate than \inda{}. Conversely, random crop is selected on Chaser (Figure~\ref{fig:ucb_chaser}). As the result, we can automatically select \inda{} or \exda{} appropriately for each task. 

\paragraph{Fully exploitation of augmentation}
In Table~\ref{table:full_exploit}, when both random color and random crop are used to improve generalization on both {\it test-bg} and {\it test-lv}, we report numerical evaluation
of UCB-\inda{} and UCB-\exda{} with other baselines on train and test tasks. Decreased train performance of DrAC and \inda{} compare to PPO show the difficulty of simultaneous training with several augmentations. Train performance of UCB-\inda{} and UCB-DrAC are improved by adaptive selecting, especially, UCB-\inda{} is better than UCB-DrAC. The gap is made due to the robustness about the change of augmentation during training. In terms of generalization, UCB-\exda{} clearly surpasses UCB-\inda{} thanks to \exda{} 
to extract all the priors from the complete set of data augmentations at the end of RL training.

\vspace{-1em}
\section{Discussion} 
\label{sec:discussion}
\vspace{-0.5em}
We have identified two most effective yet simple timings
(\inda{} and \exda{}) of data augmentation for RL,
and proposed UCB-\exda{} framework to adaptively select
the best scheduling augmentations.
We note that the effectiveness of this framework is 
restricted but specialized for RL
with the unique non-stationary nature.
Indeed, in SL without shift of data distribution and objective, it is sufficient to apply data augmentation
at the beginning \cite{golatkar2019time}.
Our framework employs the most basic multi-armed bandit
algorithm with a finite set of data augmentation.
It is interesting to investigate 
a room to improve by further optimizing continuous parameters of data augmentation for RL, c.f.,
an auto-augmentation technique to optimize continuous augmentation parameter per sample for SL \cite{gudovskiy2021autodo}.
Another promising direction is
to accelerate the distillation process of \da{}
by data condensation with augmentation
\cite{zhao2021dataset}.
This is possible with
our framework clearly separating between RL training and distillation,
and may be particularly useful to train 
distributed RL agents since a condensed data for an agent's distillation is usable for the other. 

\acksection{}

We thank Kimin Lee for helpful discussions. 
This work was supported by Institute of Information {\&} communications Technology Planning {\&} Evaluation (IITP) grant funded by the Korea government (MSIT) (No.2019-0-01906, Artificial Intelligence Graduate School Program (POSTECH)) and Institute of Information {\&} communications Technology Planning {\&} Evaluation (IITP) grant funded by the Korea government (MSIT) (No.2021-0-02068, Artificial Intelligence Innovation Hub) and 
the National Research Foundation of Korea (NRF) grant funded by the Korea government (MSIT) (No. 2021M3E5D2A01023887).

\bibliography{egbib}
\bibliographystyle{plain}

\newpage
\appendix
\section{Modified Procgen Environments}

\begin{figure}[ht]
        \subfigure[train (partial)]{
        
        \includegraphics[width=.25\columnwidth]{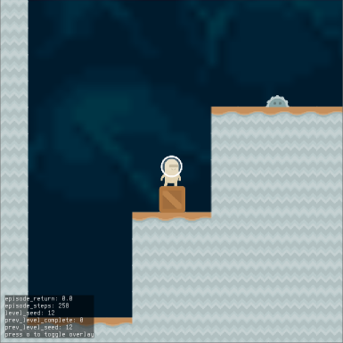}
        
        \label{fig:train_partial}

        }
        \hfill
        \subfigure[test-bg (partial)]{
        
        \includegraphics[width=.25\columnwidth]{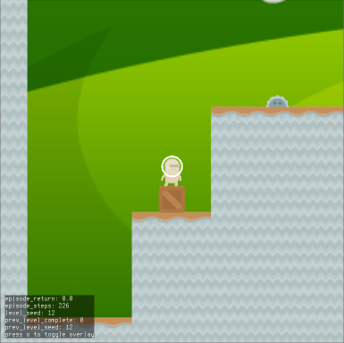}
        
        \label{fig:test_bg_partial}
        
        }
        \hfill
        \subfigure[test-lv (partial)]{
        
        \includegraphics[width=.25\columnwidth]{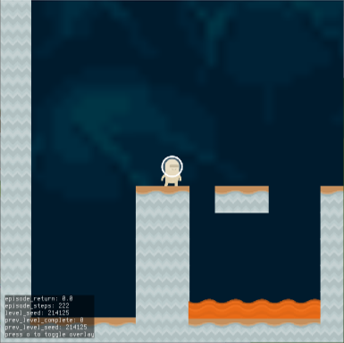}
        
        \label{fig:test_lv_partial}
        
        }
        
        \subfigure[train (fully)]{
        
        \includegraphics[width=.25\columnwidth]{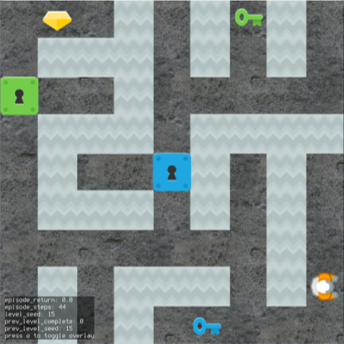}
        
        \label{fig:train_fully}

        }
        \hfill
        \subfigure[test-bg (fully)]{
        
        \includegraphics[width=.25\columnwidth]{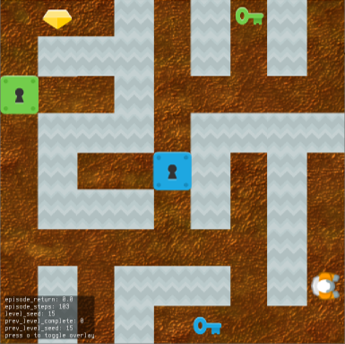}
        
        \label{fig:test_bg_fully}
        
        }
        \hfill
        \subfigure[test-lv (fully)]{
        
        \includegraphics[width=.25\columnwidth]{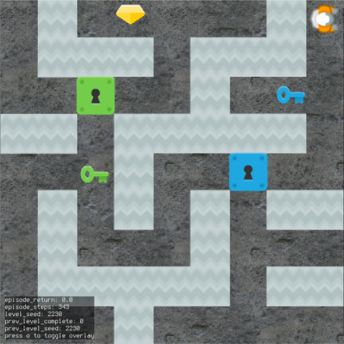}
        
        \label{fig:test_lv_fully}
        
        }
        
\caption{An example set of training and testing environments in Procgen benchmark: (upper row) an example of partially observable environment with Coinrun; (lower row) an example of fully observable environment with Heist; (left column) train: a set of levels and backgrounds for training; (center column) test-bg: the same training levels on unseen backgrounds; (right column) test-lv: a set of unseen levels on the same training backgrounds}
\label{fig:env}
\end{figure}

This section explains Modified Procgen Environments, which is designed to verify different types of generalization, backgrounds, and levels. Open AI Procgen environments \cite{cobbe2020leveraging} share background themes such as {\it space\_backgrounds}, {\it platform\_backgrounds}, {\it topdown\_backgrounds}, {\it water\_backgrounds}, {\it water\_surface\_backgrounds}.  We create new difficulties as {\it Easybg}, {\it Easybg-test}, {\it Easy-test}. {\it Easybg} generates environments which contain only one for each background, wall and agent theme. {\it Easybg-test} and {\it Easy-test} are for test about background change after trained on {\it Easybg} and {\it Easy}. Wall theme in (Climber, Coinrun, Jumper, Ninja) and Agent theme in (Climber, Coinrun) also compose with only one image resource in {\it Easybg}. Figure~\ref{fig:env} presents an example set of modes that we use in evaluation. Furthermore, We fix the exit\_wall\_choice and enemy theme in Dodgeball. We describe the usage themes in each environment, which are grouped by backgrounds theme as below:

\begin{itemize}
    \item {\it space\_backgrounds} (Bossfight, Starpilot) \\
     Background: "space\_backgrounds/deep\_space\_01.png"
    \item {\it platform\_backgrounds} (Caveflyer, Climber, Coinrun, Jumper, Miner, Ninja) \\
     Background:"platform\_backgrounds/alien\_bg.png", Coinrun (Agent color: Beige, Wall themes: Dirt), Climber (Agent color: Blue, Wall themes: tileBlue), Jumper (Wall theme: tileBlue), Ninja (Wall theme: bricksGrey)
    \item {\it topdown\_backgrounds} (Chaser, Dodgeball, Fruitbot, Heist, Leaper, Maze)\\
     Background:"topdown\_backgrounds/floortiles.png", Dodgeball (Enemy theme: "misc\_assets/character1.png", Exit\_wall\_choice: 0)
    \item {\it water\_backgrounds} (Bigfish)\\
     Background:"water\_backgrounds/water1.png"
    \item {\it water\_surface\_backgrounds} (Plunder)\\
     Background:"water\_backgrounds/water1.png"
\end{itemize}

{\it Easybg-test} uses backgrounds in each background group, except the one used in {\it Easybg}. {\it Easy-test} is only defined for Climber, Jumper, Ninja, and they compose with {\it topdown\_backgrounds}.

\section{Implementation details}
In this section, we explain about \inda{}, \exda{}, UCB-\inda{} and other baselines. We train the agent with IMPALA-CNN \cite{espeholt2018impala} in every experiment. 
\subsection{\inda{}} 
We use PPO \cite{schulman2017proximal} as a base RL algorithm, For data efficiency, we store the observations during RL training in buffer $\set{D}_O$. Before \da{} phase, we also make policy buffer  $\set{D}_\Pi$, value function buffer $\set{D}_V$ and augmented observation buffer $\set{D}_\phi$ for distillation, because we only use one network model. We randomly sample pairs of $(o, \pi, V)$ from buffer, and minimize loss function $L_{\da{}}(\theta)$.  We reuse the sample three times like PPO, it can be controlled by  $\#$ Epochs of \da{}. We did a greed searches for learning rate of \da{} $l_{DA} \in [ 1\times 10^{-3}, 5
 \times 10^{-4}, 2 \times 10^{-4}, 1 \times 10^{-4}, 5 \times 10^{-5} ] $ and interval $I \in [1, 5, 10 ]$ and found the best combination $l_{\da{}} = 10^{-4}$ and interval $I =5$. We fix the buffer size $\set{D}_O = 40960$, because we collect the observations during five RL phases $(5 \times 256 \times 32)$. We describe the every hyperparameter as below:

\begin{table}[H]
\centering
    \begin{tabular}{cc}
    \toprule
    Hyperparamter & Value  \\
    \midrule  
        $\gamma$ & 0.999 \\
        $\lambda$ & 0.95 \\
        \# Timesteps per rollout & 256 \\
        \# Epochs per rollout & 3 \\
        \# Minibatches per epoch & 8\\
        Reward Normalization & Yes \\
        \# Workers & 1\\
        \# Environments per worker & 32 \\
        Total timesteps & 25M \\
        LSTM & No\\
        Frame Stack & No \\
        Optimizer & Adam optimizer   \\
        Entropy bonus & 0.01\\
        PPO clip range & 0.2 \\
        Learning rate & $5 \times 10^{-4}$  \\
        Interval $I$ & 5\\
        Size of $\set{D}_O$ & 40960 \\
        \# Epochs of \da{} & 3 \\
        Learning rate of DA $l_{\da{}}$ & $1\times10^{-4}$\\ 
        Image transformation $\phi$ &  Any augmentation\\
    \bottomrule
\end{tabular}
\end{table}

\subsection{\exda{}}
In \exda{}, we generate and store $(o, \pi, V)$ using $f_{\theta_{old}}$ in buffer $\set{D}$. The optimal buffer size depends on the episode length of each environment. However, we standardize the buffer size as 0.5M in every environment. We augment the observation with three epochs intervals when using randomized augmentation methods. We did greed searches for \# minibatches $ [1024, 2048, 4096 ] $ and learning rate $[  5\times10^{-4}, 1\times 10^{-3}, 2\times 10^{-3} ]$. As a result, we select $\#$ of minibatches $4096$ and a learning rate $1e-3$.We describe every hyperparameter as below:
\begin{center}
    \begin{tabular}{cc}
    \toprule
    Hyperparameter & Value  \\
    \midrule  
        Size of $\set{D}_O$ & 0.5M \\
        \# Epoch & 30 \\
        \# Minibatches per epoch & 4096\\
        Learning rate & $1\times 10^{-3}$  \\
        \# Workers & 1\\
        Optimizer & Adam optimizer   \\
        Image transformation $\phi$ &  Any augmentation\\
    \bottomrule
\end{tabular}
\end{center}

\subsection{UCB-InDA}

We use UCB-\inda{} as a discriminator to determine the necessity of augmentation during the training. The gain of an augmentation is a mean of return during interval I, $G(s) = \frac{1}{I}\sum_{i=0}^{j-1} R(s+j)$. The return is computed by the sum of estimated advantage and predicted value, which are expected value of the agent trajectory, $R(s) = \hat{\mathbb{E}}_{(o_t,a_t) \sim \pi_\theta} [\hat{A}_t + V_\theta(o_t)]$. The $\hat{A}_t$ is advantage from Generalized Advantage Estimator \cite{schulman2015high}. Thus, we can evaluate how augmentation affects the return on the agent trajectory. However, the distribution of return is non-stationary, as the agent policy is changed. Therefore, we use the window average gain $\bar{G_\phi}$ rather than the whole gain from the past evaluation. Furthermore, the drastic change of return causes the gap of gain between the augmentation according to sampling time at the transient time of training and leads to poor exploration about some augmentation methods. For stable exploration, we fix the minimum exploration frequency and use forced exploration method after the minimum exploration as below: 
\begin{equation}
    \bar{G}_{\phi_{max}}(s) + c \sqrt{\frac{\log(s)}{N_{\phi_{max}}(s)}} \leq \bar{G}_{\phi_{min}}(s) + c \sqrt{\frac{\log(s)}{N_{\phi_{min}}(s)}}
\end{equation}
\begin{equation}
    c = \frac{\bar{G}_{\phi_{max}} - \bar{G}_{\phi_{min}} + \epsilon}{\sqrt{\log(s)} \times \max(\frac{1}{\sqrt{N_{\phi_{min}}(s)}} - \frac{1}{\sqrt{N_{\phi_{max}}(s)}}, \frac{1}{\sqrt{W-1}} - \frac{1}{\sqrt{W}})}
\end{equation}
where $\phi_{max} = \argmax_{\phi \in \Phi} \bar{G}_\phi,  \phi_{min} = \argmin_{\phi \in \Phi} \bar{G}_\phi$.
We set the hyperparameter as below table:
\begin{center}
    \begin{tabular}{cc}
    \toprule
    Hyperparameter & Value  \\
    \midrule  
       Window size of gain W & 3 \\
       Minimum exploration frequency & 15 \\
    \bottomrule
\end{tabular}
\end{center}

\subsection{Baselines}
We compare \exda{} and \inda{} with PPO \cite{cobbe2020leveraging}, DrAC \cite{raileanu2020automatic}, Rand-FM \cite{lee2019network}, RAD \cite{laskin2020reinforcement}. Every baseline is based on PPO \cite{cobbe2020leveraging} and we adopt the implementation of PPO in \cite{cobbe2020leveraging}. 
\vspace{-1em}
\begin{itemize}
    \item DrAC \cite{raileanu2020automatic} regularizes both policy and value function as self-supervised learning. Regularization term have hyperparameter $\alpha_r$ for ratio with PPO objective. We use the hyperparameter recommended by the author. \\
    \item Rand-FM \cite{lee2019network} is composed with random convolution networks and feature matching. They also need hyperparameter $\beta$ for ratio between feature matching and PPO objective. We use same $\beta$ with author. \\
    \item RAD \cite{laskin2020reinforcement} naively use augmented observations in state distribution. Thus, there are no additional hyperparameters. 
\end{itemize}

We describe the hyperparameter of baselines in below table:
\begin{center}
    \begin{tabular}{cc}

    \toprule
    Hyperparameter & Value  \\
    \midrule  
        $\gamma$ & 0.999 \\
        $\lambda$ & 0.95 \\
        \# of timesteps per rollout & 256 \\
        \# of epochs per rollout & 3 \\
        \# of Minibatches per epoch & 8\\
        Reward Normalization & Yes \\
        \# of Workers & 1\\
        \# of environments per worker & 64 \\
        Total timesteps & 25M \\
        LSTM & No\\
        Frame Stack & No \\
        Optimizer & Adam optimizer   \\
        Entropy bonus & 0.01\\
        PPO clip range & 0.2 \\
        Learning rate & $5 \times 10^{-4}$  \\
        $ \alpha_r $ (DrAC) & 0.1 \\
        $\beta$ (Rand-FM) & 0.002 \\ 
    \bottomrule
\end{tabular}
\end{center}

\section{Data augmentation}
In our experiments, we use five augmentation methods: {\it crop, grayscale, cutout color, random convolution} and {\it color jitter}. We refer the implementation of augmentations from Lee {\it et al.} \cite{lee2019network} ({\it random convolution}), Laskin {\it et al.} \cite{laskin2020reinforcement} ({\it cutout color, color jitter})  and Raileanu {\it et al.} \cite{raileanu2020automatic} ({\it grayscale, crop}). We expect the generalization about background change from {\it random convolution, color jitter, gray, cutout color}. About the change of levels, we use {\it crop} and {\it cutout color} for generalization. Examples of data augmentation are represented below:
\begin{figure}[H]
    \centering
    \subfigure[Original]{
        \includegraphics[width=0.2\textwidth]{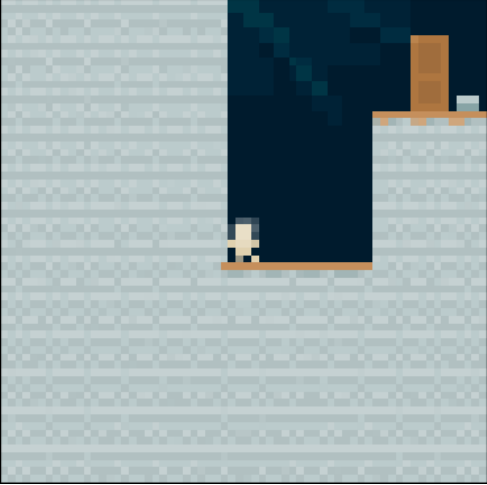}
      
        }
    \subfigure[Rand Conv]{
        \includegraphics[width=0.2\textwidth]{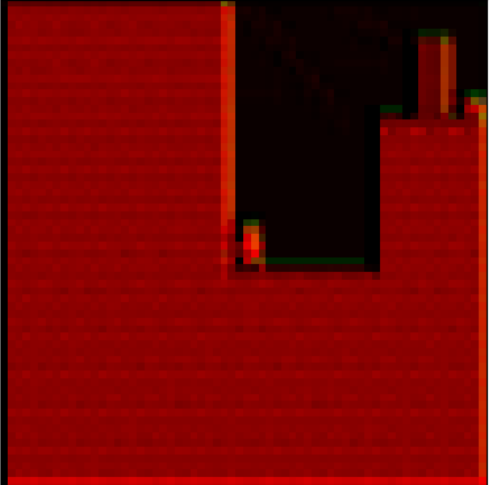}
       
        }   
    \subfigure[Crop]{
        \includegraphics[width=0.2\textwidth]{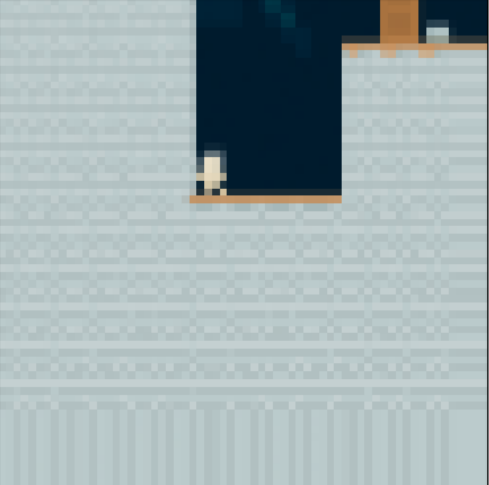}
        } \\
    \subfigure[Color jitter]{
        \includegraphics[width=0.2\textwidth]{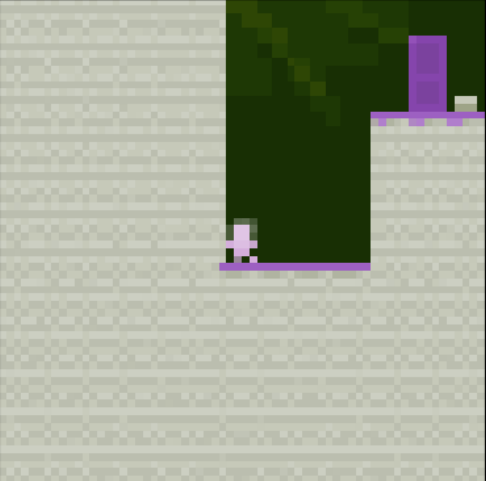}
        }
    \subfigure[Gray]{
        \includegraphics[width=0.2\textwidth]{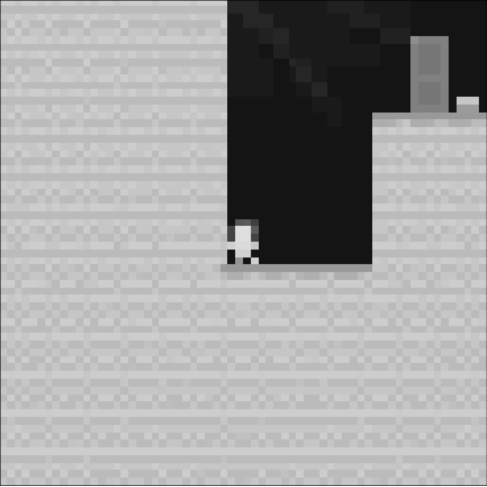}
        }   
    \subfigure[Cutout color]{
        \includegraphics[width=0.2\textwidth]{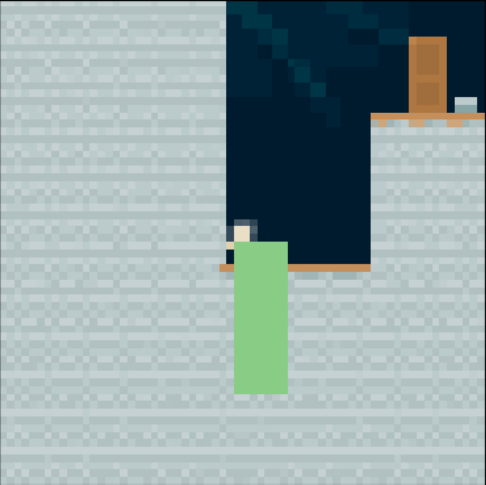}
        }
    \caption{Examples of visual augmentations}
    \label{fig:augmentation}
\end{figure}

\begin{figure}[H]
    \includegraphics[width=0.8\linewidth]{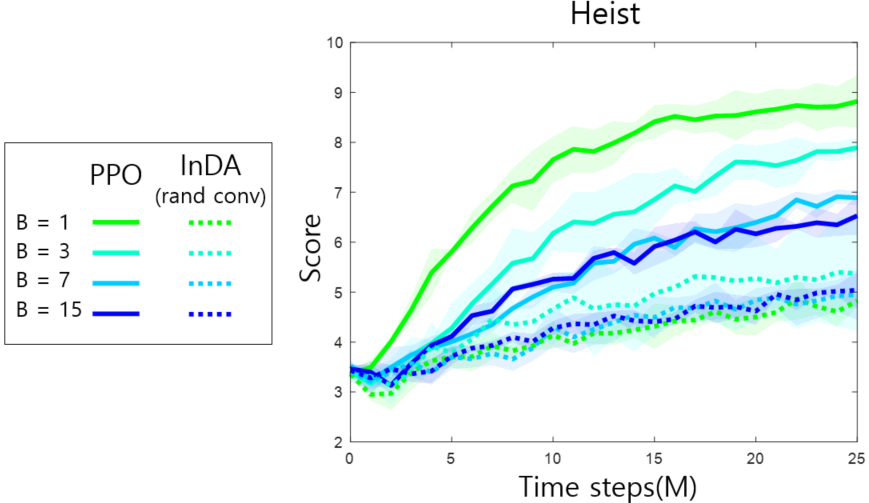}
    \caption{{\it Interference from increased complexity of learning.}
    We present train performance when training agent on
    Heist({\it easybg}-$B$), where 
    {\it easybg}-$B$ is a variant of 
    {\it easybg} mode with $B$ backgrounds.
    We plot the train performance curves of PPO and InDA with random convolution over training epochs in absolute score.
    }
    \label{fig:bg_diversity}
\end{figure}

\section{Interference from increased complexity of learning}
Even if the policy for the original observation is fixed and augmentation is used, learning may still be hindered. In Figure~\ref{fig:bg_diversity}, we report training curves of the agents on train tasks
of different degree of background diversity.
\inda{} with random convolution is anticipated to lead the prior on color consistency,
which seems helpful to handle background diversity.
However, PPO outperforms \inda{} 
in terms of the sample complexity to master train tasks, while
the gap is decreasing as the background diversity increases.
This implies that diverse backgrounds make hard to train by increased complexity of learning, and also data augmentation can cause similar difficulty even with right prior, when using it during RL training.
We further remark that \exda{} using random convolution
after PPO trained on
Heist({\it easybg-1}) achieves 8.15 on Heist({\it easybg-15}),
which is much higher than PPO's 6.53 trained on Heist({\it easybg-15}).
This suggests the importance of simplifying the train task
and the utility of \exda{} which completely separates 
RL training and distillation with augmentation.

\section{Robustness in loss function change}
\begin{figure}[H]
    \centering
    \includegraphics[width=\textwidth]{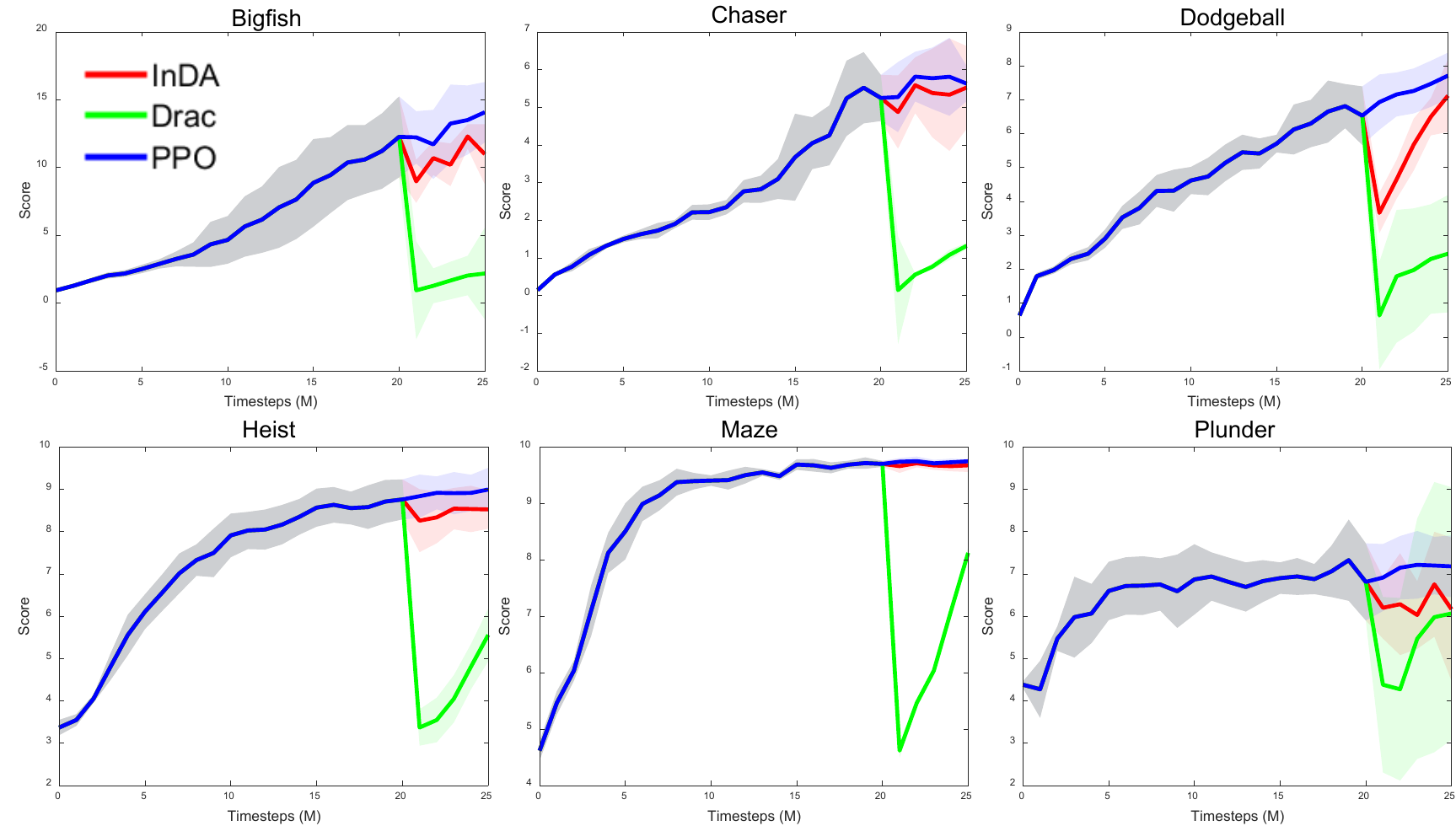}
    \caption{Comparison of the training performance when {\it crop} is applied after 20M timesteps with InDA and Drac. }
    \label{fig:20M+crop_train}
\end{figure}

\begin{figure}[H]
    \centering
    \includegraphics[width=\textwidth]{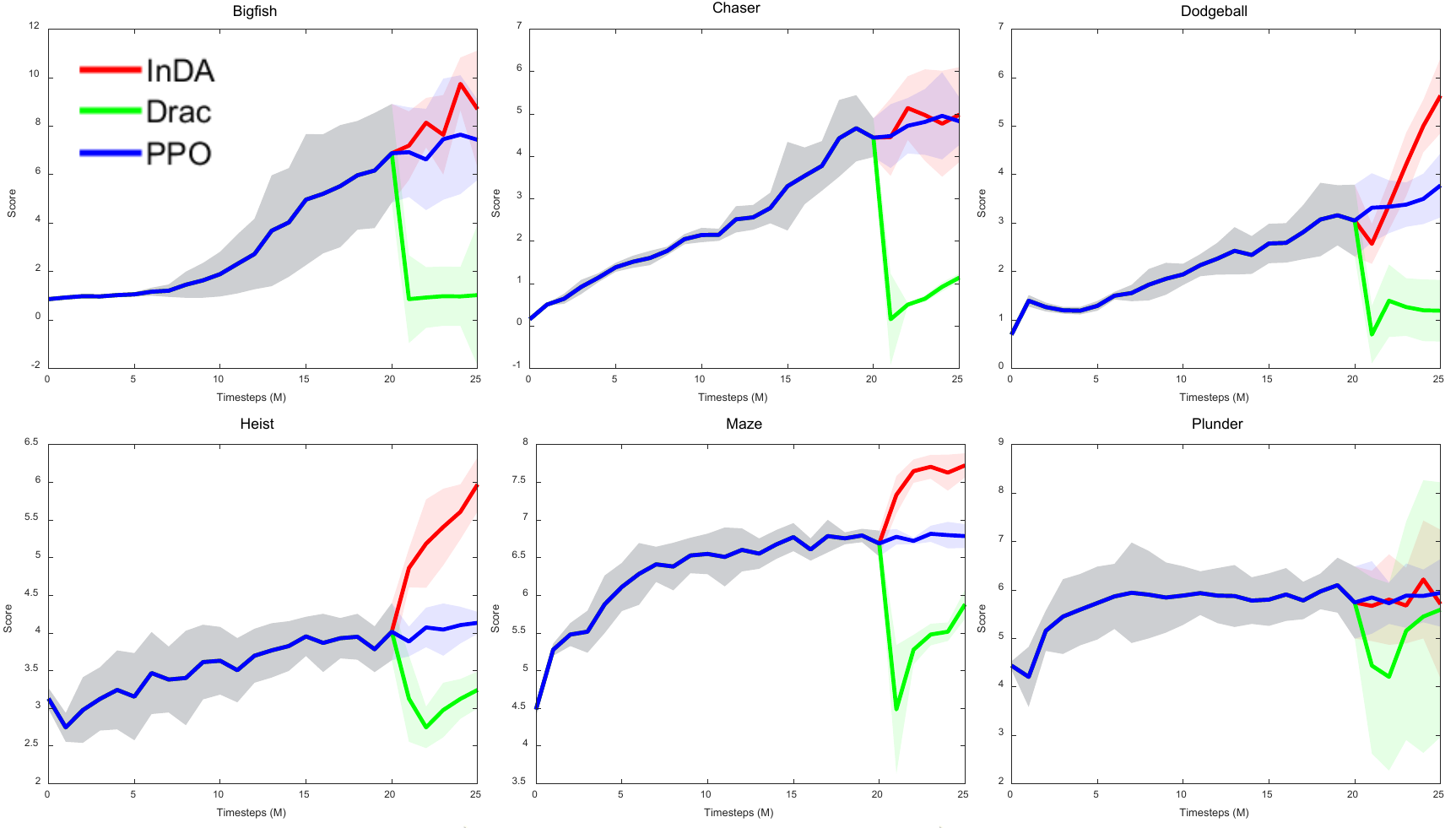}
    \caption{Comparison of the test performance when {\it crop} is applied after 20M timesteps with InDA and Drac.}
    \label{fig:20M+crop_test}
\end{figure}

In \exda{}, we transfer the policy after training 20M time steps with PPO. Thus, we explain why other augmentations are not used after pre-training. We compare the results of training and test performance with Drac \cite{raileanu2020automatic}, Rand-FM \cite{lee2019network}, Rad \cite{laskin2020reinforcement} when we train each method for 5M after training PPO for 20M time steps. We use random convolution and crop as data augmentation methods, and we do not compare with RAD when we use crop in Figure~\ref{fig:20M+crop_train} and Figure~\ref{fig:20M+crop_test}. The {\it crop} method used in our paper do not work well in RAD, because they use a different crop method with \cite{raileanu2020automatic} in their paper \cite{laskin2020reinforcement}. \inda{} is more stable than others in training, and it affects generalization performance.

Every training curves decline immediately after starting training with augmented observations at 20M time steps. The objective function is changed to each baseline, and augmented data is newly added to data distribution. Thus, the optimizer should find a new optimal point for new objective function and data. During find the new optimal points, the agent learns along with the different directions from the optimization direction in pure PPO. Thus, performance can be degraded because the learning direction on loss landscape is different from maximizing rewards on non-augmented data in PPO. In spite of using self-supervised learning or representation learning, the policy is changed because they update the same network's parameter for matching policy or latent features, such as DrAC \cite{raileanu2020automatic} and Rand-FM \cite{lee2019network}. However, \inda{} is more stable than the others because we distill the fixed policy and value using DA. It does the stable training through conserving the policy on non-augmented observations during optimizing for augmented data. 

\begin{figure}[H]
    \centering
    \includegraphics[width=\textwidth]{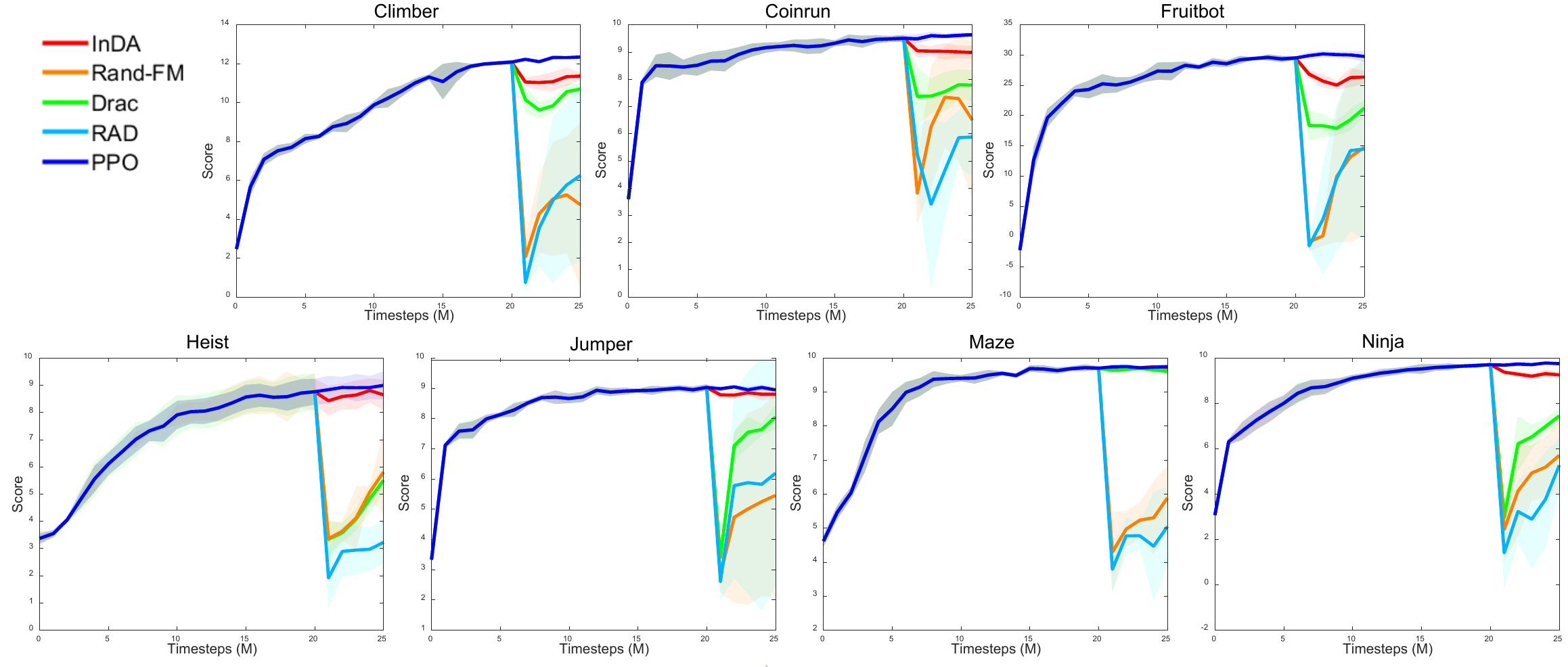}
    \caption{Comparison of the training performance when {\it random convolution} is applied after 20M timesteps with various augmentation methods. }
    
\end{figure}

\begin{figure}[H]
    \centering
    \includegraphics[width=\textwidth]{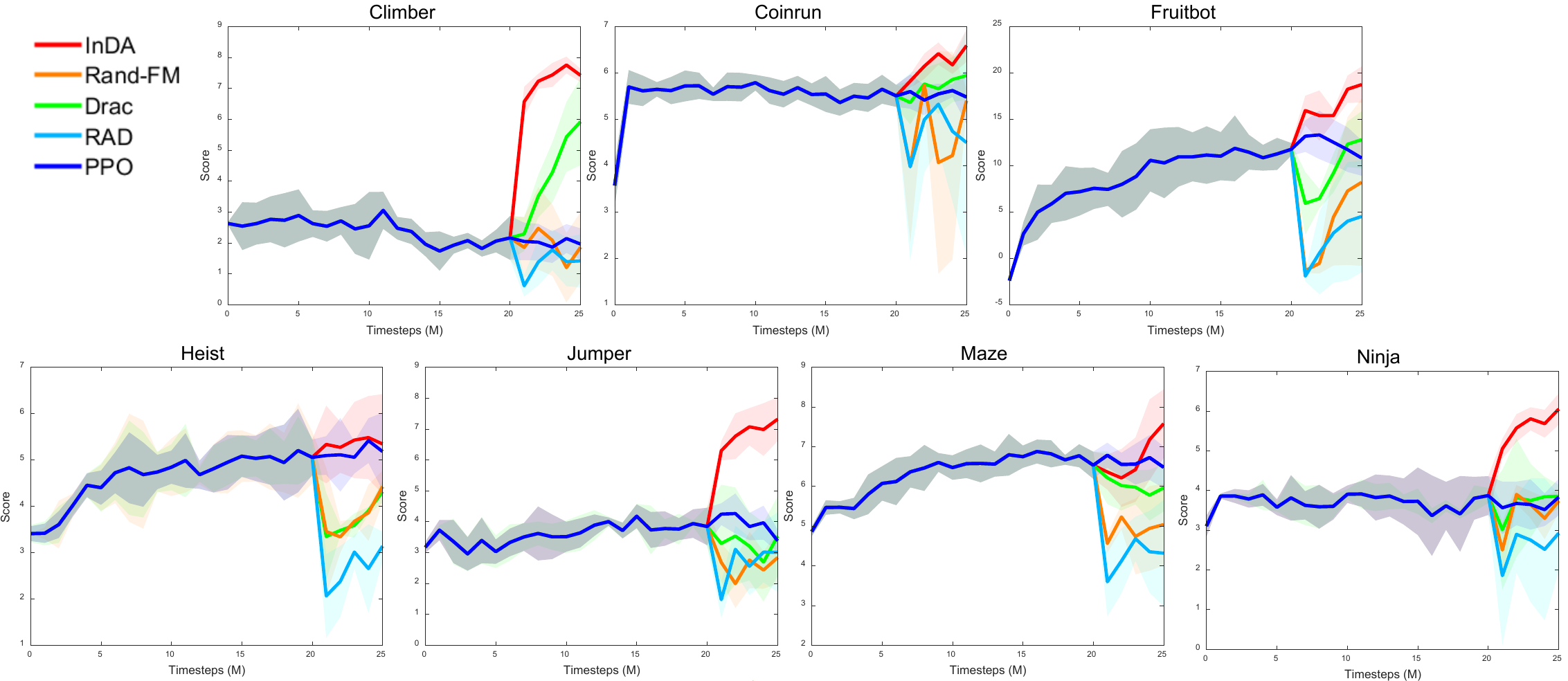}
    \caption{Comparison of the test performance when {\it random convolution} is applied after 20M timesteps with various augmentation methods. }
    
\end{figure}

\section{Robustness against wrong augmentation}
In this section, we verify the robustness of \exda{} from an obvious wrong augmentation: just making an black image, denoted by {\it black}. Using this clearly interferes with RL training. We compare PPO, \inda{}, \exda{}, UCB-\inda{}, UCB-\exda{} and DrAC. The hyperparameter is almost the same with Section~\ref{sec:eval} except we use {\it black} instead of {\it random color} and {\it random crop}. For UCB-based methods, we use 4 arms:  {\it black}, {\it random color}, {\it random crop}, and {\it no aug}. As expected, \exda{} preserves the PPO score, but \inda{} and DrAC degrade the score. Furthermore, UCB-\inda{} improves the performance in Chaser, and also it almost preserves the score in Heist. Lastly, UCB-\exda{} also maintains the score from UCB-\inda{}.

\begin{table*}[ht]
\tiny
    \centering
    \renewcommand{\multirowsetup}{\centering}
    \resizebox{\textwidth}{!}{
      \begin{tabular}{ c|ccc|ccc}
        \toprule
        \multirow{1}{5mm}{Env} & PPO & DrAC   & InDA  & ExDA  & UCB-InDA & UCB-ExDA  \\

        \midrule  
        \multirow{1}{5mm}{Heist}    & \bf{9.2  $\pm$ 0.46}  & 7.35 $\pm$ 0.684 & 6.72 $\pm$0.419  & 8.6 $\pm$ 0.189 & 8.84  $\pm$ 0.307 & 8.64$\pm$ 0.264  \\

        \midrule  
        \multirow{1}{5mm}{Chaser} & 5.63$\pm$ 1.12 & 1.49$\pm$0.036   & 5.43$\pm$ 0.633 & 5.1$\pm$0.331 &  \bf{6.74}$\pm $0.588 & 6.28$\pm$ 0.436  \\

        \bottomrule
      \end{tabular}}
    \caption{Robustness from the wrong augmentation }
    \label{table:wrong_aug} 
\end{table*}

\section{Ablation study of \exda{}}
\subsection{Initialization and regularization term}
\begin{figure}[H]
    \centering
    \includegraphics[width=\textwidth]{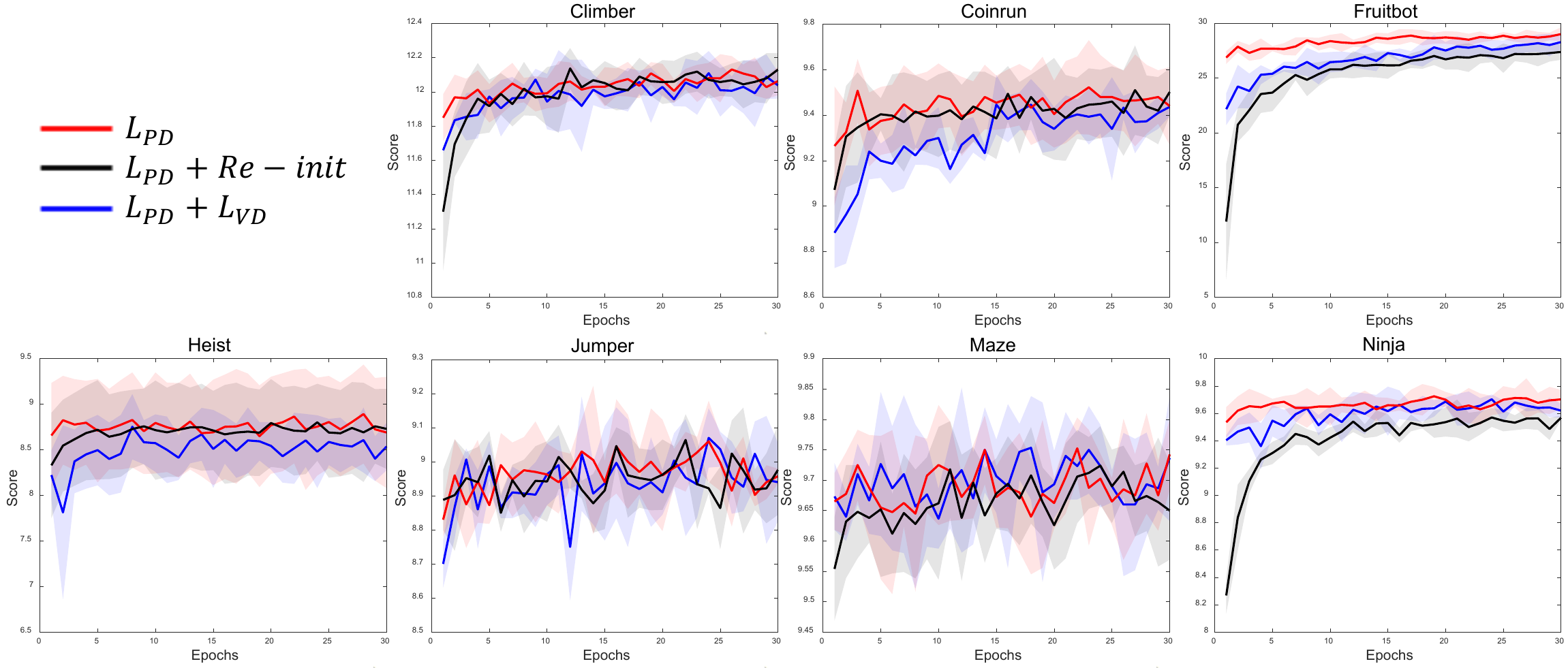}
    \caption{Training performance of \exda{} with re-initialization or regularization with value funtion.}
    \label{fig:abl-exda-train}
\end{figure}

\begin{figure}[H]
    \centering
    \includegraphics[width=\textwidth]{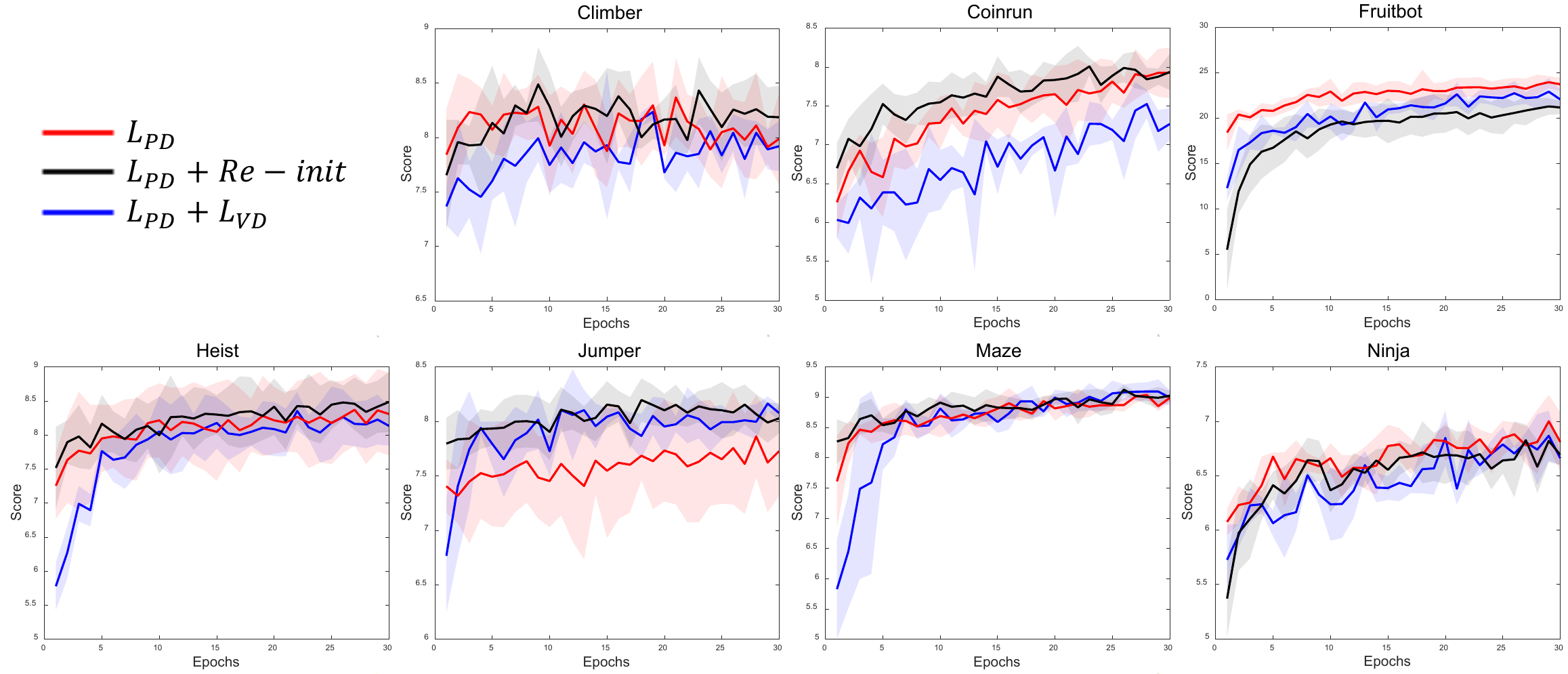}
    \caption{Test performance of \exda{} with re-initialization or regularization with value funtion on unseen backgrounds. }
    \label{fig:abl-exda-test}
    
\end{figure}

In this section, we do an ablation study about the factor of \exda{}. We mention the loss function and re-initialization issue in \autoref{sec:exda}. \exda{} does not have to minimize $L_{VD}$ because the value function is useless after RL training. The below results show that $L_{VD}$ cannot give any benefit in \exda{}. Thus, we only use $L_{PD}$ for computational complexity. Furthermore, we also compare to verify the effect of non-stationarity with a re-initialized agent before distillation. Igl {\it et al.} \cite{igl2020impact} argued that the non-stationarity causes the reduction of generalization.  However, the re-initialization is not critical in test performance, as shown in Figure~\ref{fig:abl-exda-test}. Moreover, sometimes re-initialization makes it difficult to distill training performance such as Fruitbot and Ninja in Figure~\ref{fig:abl-exda-train}. We use {\it random convolution } as an augmentation method in here.    

\subsection{\exda{} after \inda{} with various backgrounds}
When augmentation helps the training, \exda{} struggle to follow the training performance of \inda{} because \exda{}'s training performance is limited by pre-trained agent's policy. Thus, we use \inda{} for \exda{}'s pre-training , and call it as \exda{} (\inda{}). As shown in \autoref{table:ExDA (InDA)-test}, \exda{} (\inda{}) is comparable to \inda{}, but not beyond. Thus, unless there is a meaningful difference in training performance, \exda{} has no better generalization than \inda{}. However, in computational complexity, \exda{} is more efficient than others such as \inda{} and DrAC when they have a similar performance. In the following section, we discuss computational complexity. 


\begin{table}[H]
\centering
\begin{tabular}{cccccccc}
\toprule
\multicolumn{1}{c}{Easy} & PPO & InDA & ExDA (PPO) & ExDA (PPO) + reinit & ExDA (InDA) & ExDA (InDA) + reinit \\
\midrule
\multirow{2}{*}{Jumper}  & 8.55 & 8.94 & 8.5   & 8.6   & 8.83  & 8.83  \\
                         & \std0.17 & \std0.09 & \std0.183 & \std0.156 & \std0.215 & \std0.126 \\
\multirow{2}{*}{Ninja}   & 7.49 & 8.88 & 7.03  & 7.23  & 8.71  & 8.56  \\
                         & \std0.42 & \std0.34 & \std0.058 & \std0.159 & \std0.344 & \std0.394 \\
\multirow{2}{*}{Climber} & 8.63 & 8.5  & 8.1   & 8.09  & 8.16  & 7.99  \\
                         & \std0.46 & \std0.29 & \std0.268 & \std0.268 & \std0.441 & \std0.383 \\
                         \bottomrule
                         
\end{tabular}

\caption{The comparison with diverse agents which are trained with \exda{} }
\label{table:ExDA (InDA)-train}
\end{table}
\vspace{-1em}
\begin{table}[H]
\centering
\begin{tabular}{cccccccc}
\toprule
\multicolumn{1}{c}{Easy} & PPO & InDA & ExDA (PPO) & ExDA (PPO) + reinit & ExDA (InDA) & ExDA (InDA) + reinit \\
\midrule
\multirow{2}{*}{Jumper}  & 6.85 & 7.94 & 7.54  & 7.48  & 7.98  & 7.67  \\
                         & \std0.19 & \std0.19 & \std0.158 & \std0.154 & \std0.148 & \std0.155 \\
\multirow{2}{*}{Ninja}   & 6.29 & 6.5  & 5.56  & 5.73  & 6.27  & 5.94  \\
                         & \std0.19 & \std0.19 & \std0.158 & \std0.154 & \std0.148 & \std0.155 \\
\multirow{2}{*}{Climber} & 6.96 & 7.28 & 7.06  & 6.89  & 6.8   & 5.45  \\
                         & \std0.65 & \std0.35 & \std0.541 & \std0.237 & \std0.441 & \std0.383 \\
                         \bottomrule
\end{tabular}
\caption{Test performance of agents, which is trained on easy mode with random convolution.  }
\label{table:ExDA (InDA)-test}
\end{table}

\subsection{Computational complexity}
We compare the computational complexity with \exda{} and \inda{}. \inda{} do \da{} for every 25M observations during training and reuse the sample in three times. However, \exda{} only use 0.5M for \da{} during 30 epochs. Thus, \exda{} is almost 5 times more efficient than \inda{} by rough calculation. Furthermore, the \exda{} saves the time for augmentation compared to \inda{}. When we train with same computational setting (GPU: GeForce RTX 2080 TI), \exda{} only consumes 5 hours + 2 hours (PPO) when using random convolution, but, \inda{} consumes 18 hours. Thus, we recommend \exda{} when \inda{} cannot give a meaningful gain in training performance.

\section{Comparison between \inda{} and \exda{} using the same steps of RL training}
 In every experiment in the paper, we set the evaluation setup to be somewhat unfavorable to \exda{} (using 20M time steps of RL training followed by additional 0.5M steps of distillation; denoted by \exda{}(20M)) compared to \inda{} or other baselines (25M time steps of RL training) to clearly avoid potential complaint about the extra 0.5M steps for \exda{}, It is obvious that the performance of \exda{} is improved if we put more time steps for RL training, and thus the benefit of \exda{} compared to \inda{} becomes more conspicuous if \exda{} is the effective timing. Thus, we do an additional experiment evaluating \exda{} (25M) using 25M RL time steps and 0.5M distillation time steps as below table:

\begin{table}[H]
\centering
\begin{tabular}{ccccccccc}
\toprule
\multicolumn{1}{c}{Easybg}& & PPO & DrAC & RAD  &\inda{} & \exda{}(20M) & \exda{}(25M) \\
\midrule
\multirow{3}{*}{Heist}& Train & \bf{9} & 5.95 & 7.94  & 5.15  & 8.72  & 8.93  \\
                      & Test-bg  & 5.18 & 5.47 & 4.78  & 4.96  & 8.15  & \bf{8.26}  \\
                      & Test-lv  & 4.13 & 5.4 & 3.81  & \bf{5.91}  & 5.35  & 5.41  \\
\bottomrule
\end{tabular}
\caption{ Additional RL training before \exda{} }
\label{table:ExDA(25M)}
\end{table}

The result of \exda{}(25M) reinforces our main message: postponing data augmentation when it generates a severe interference with RL training in Table~\ref{table:ExDA(25M)}. \exda{}(25M) has a slight drop in train score compared to PPO, but it could be eventually removed if we put (slightly) more steps for distillation.

\newpage

\section{UCB with a large set of augmentation}
\begin{figure}[ht]
    \centering
    \centering
    \subfigure[Augmentation selection]{
        \includegraphics[width=0.42\textwidth]{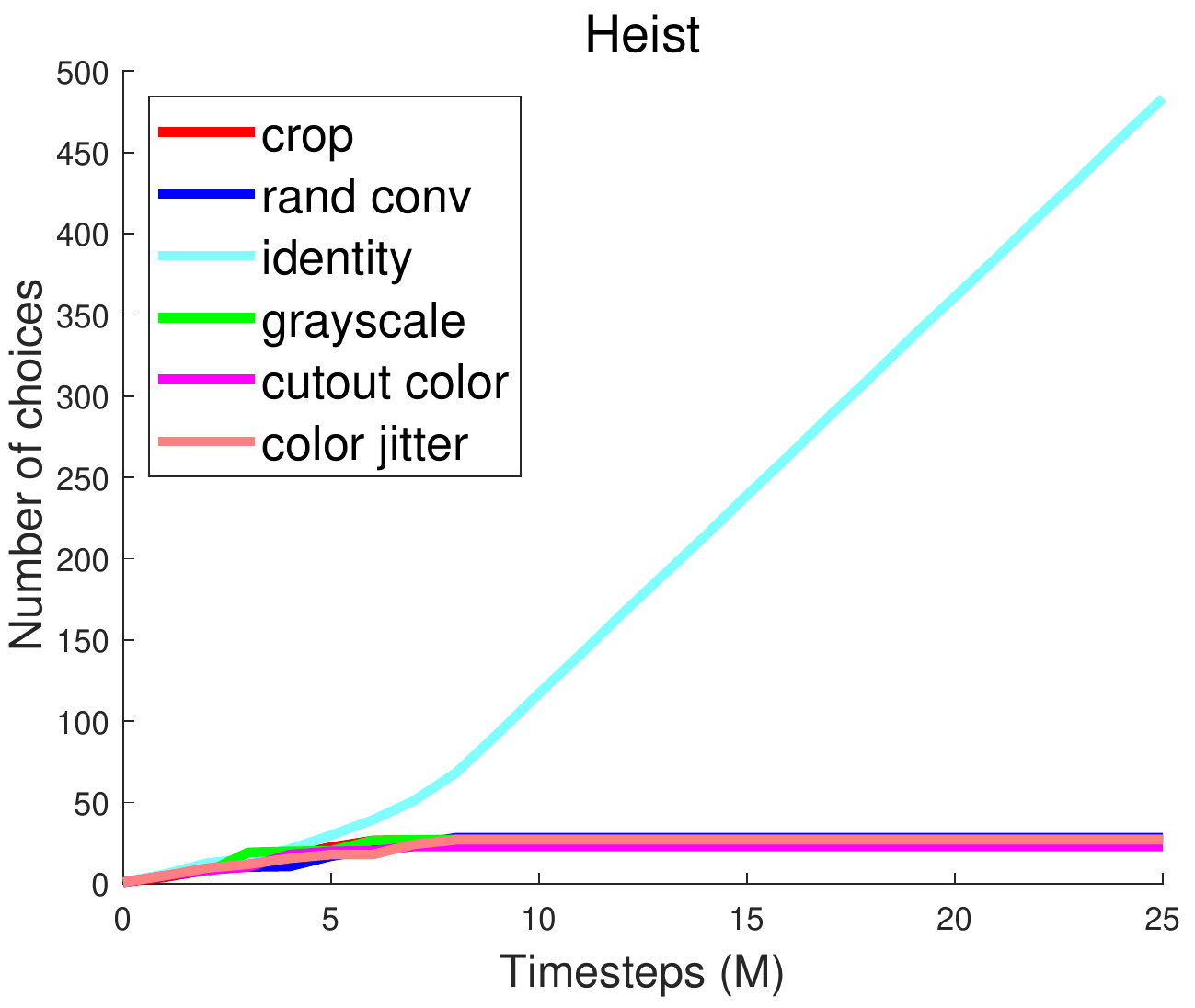}
        \label{fig:heist_aug_choice}
        }
    \subfigure[Performance]{
        \includegraphics[width=0.42\textwidth]{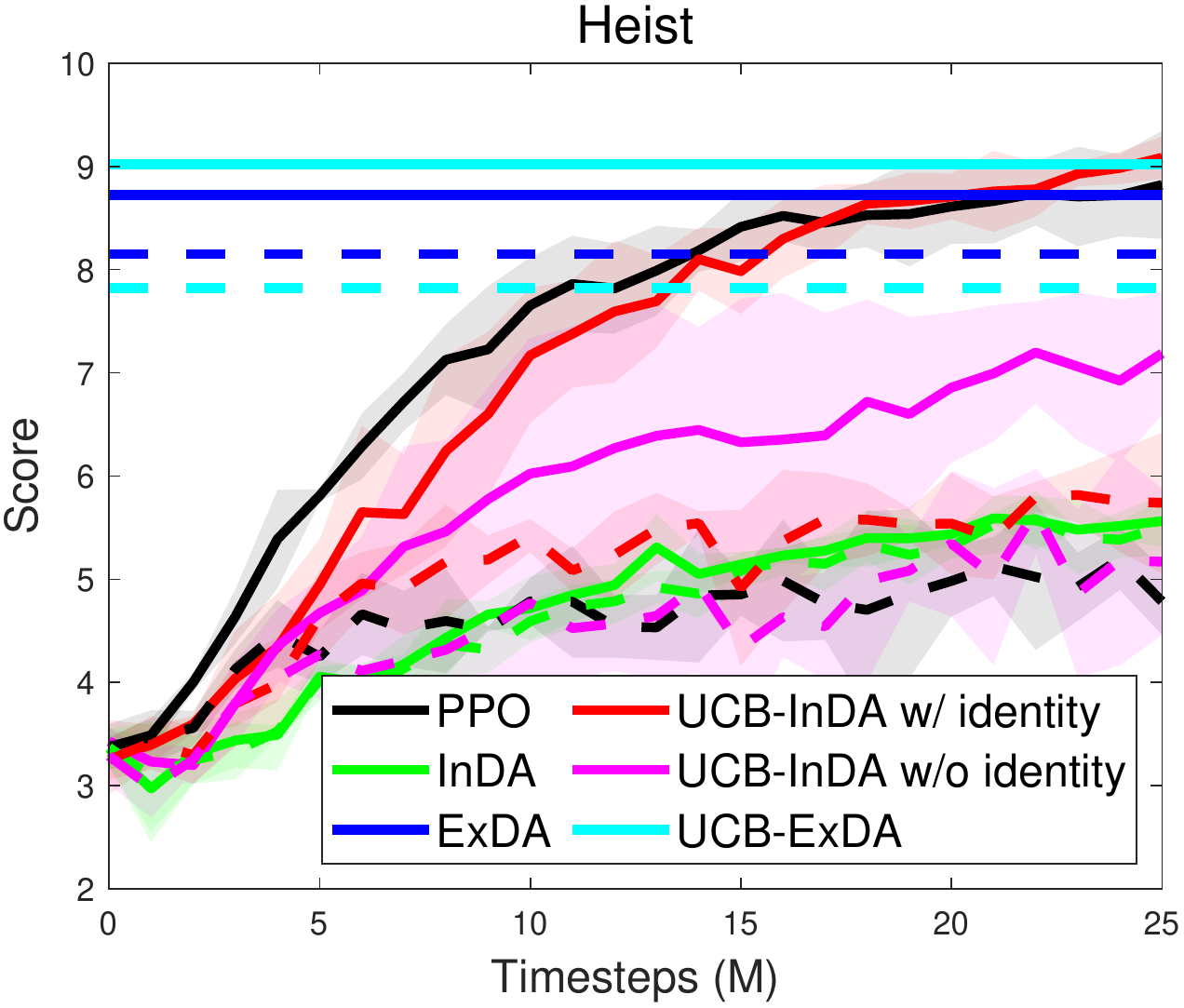}
        \label{fig:heist_ucb}
        }
    \caption{Figure~\ref{fig:heist_aug_choice} show the selected number of augmentations by UCB on Heist. We compare two UCB-\inda{}s, w/ and w/o identity function with PPO, \inda{}, \exda{}, UCB-\exda{} on Heist {\it easybg} in Figure~\ref{fig:heist_ucb}. UCB-\inda{} is trained after UCB-\inda{} w/ identity, we use {\it random convolution} as a data augmentation in \inda{}, \exda{}. Solid line: train performance; dotted line: test performance. \exda{} achieves larger test performance than \inda{} by preserving train performance. Moreover, UCB-\inda{} w/ identity outperforms UCB-\inda{} w/o identity in the training. }
    \label{fig:heist}
\end{figure}

In Figure~\ref{fig:ucb_selection}, we only use three arms such as {\it random color}, {\it random crop} and {\it no augmentation}. Thus, we try to select the usuful augmentations at each time among a large set of augmentation, which contains {\it gray}, {\it cutout color}, {\it random convolution}, {\it color jitter}, {\it random crop} and {\it no augmentation}. In Figure~\ref{fig:heist_aug_choice}, we show the result of arm selection with UCB-InDA. The {\it identity} function is most frequently used in the same as Figure~\ref{fig:ucb_selection}. On the other hand, we ablate the neccesity of identity function in UCB-\inda{}, it shows that the {\it identity} function is needed.  

\section{Time matter in training}
This section shows every result of Figure~\ref{fig:time_sensitivity} about time dependency with \inda{}. We experiment with {\it random convolution, crop, color jitter} and evaluate the test on unseen backgrounds ({\it random convolution, color jitter}) and levels ({\it random crop}). However, the effect of generalization is hard to recognize in most cases, as shown in \autoref{sec:benchmark}. Thus, we mainly discuss the most effective augmentation, such as random convolution and crop in the main paper, and only represent some environments that have helped the generalization by color jitter. {\it easybg} mode is used as default mode with three {\it easy} mode (Climber, Jumper, Ninja) in our experiments. The shaded regions and solid line represent the standard deviation and mean, across five runs.

\subsection{Random convolution}

\begin{figure}[H]
    \centering
    \includegraphics[width=0.84\textwidth]{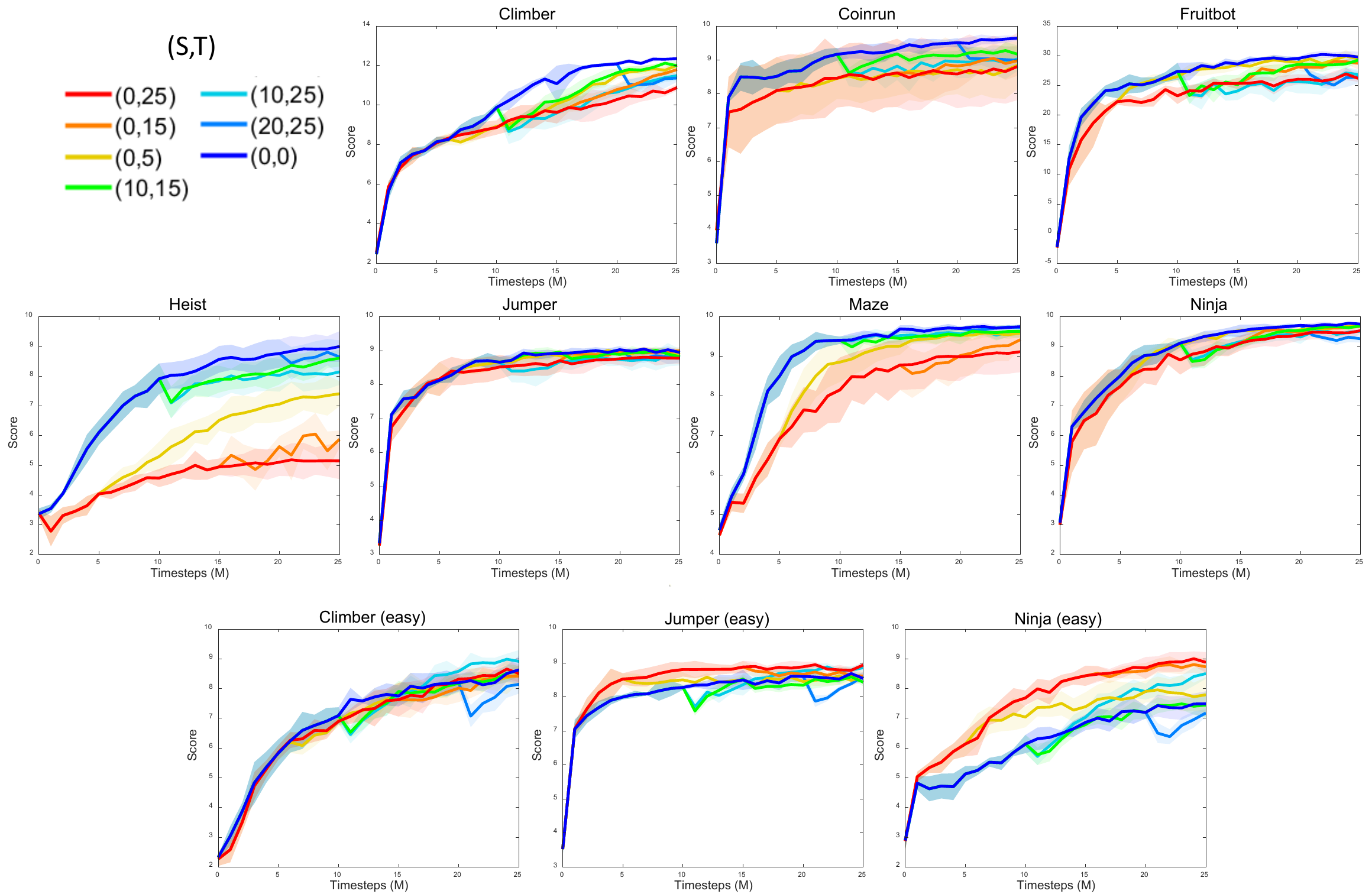}
    \caption{Comparison of training performance according to usage period of augmentation with \inda{} ({\it random convolution}): The {\it easybg} is disturbed by {\it random convolution}, but, {\it easy} mode is improved training performance by {\it random convolution}.}
    
\end{figure}

\begin{figure}[H]
    \centering
    \includegraphics[width=0.84\textwidth]{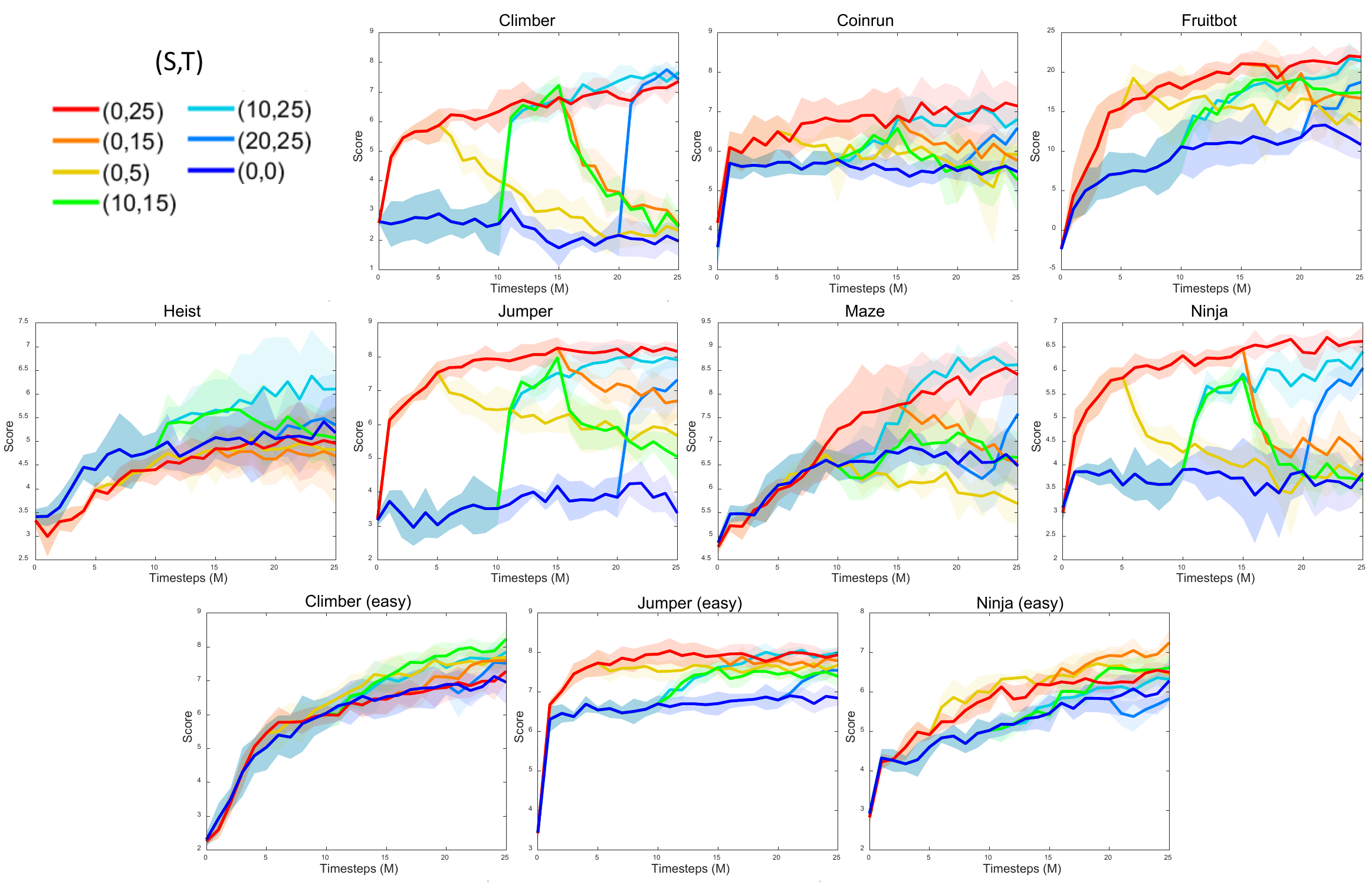}
    \caption{Comparison of generalization on unseen backgrounds according to usage period of augmentation with \inda{} ({\it random convolution}): Most cases' tendencies are coincidence with the jumper, which is mentioned in the main paper.}
    
\end{figure}

\subsection{Crop}

\begin{figure}[H]
    \centering
    \includegraphics[width=0.9\textwidth]{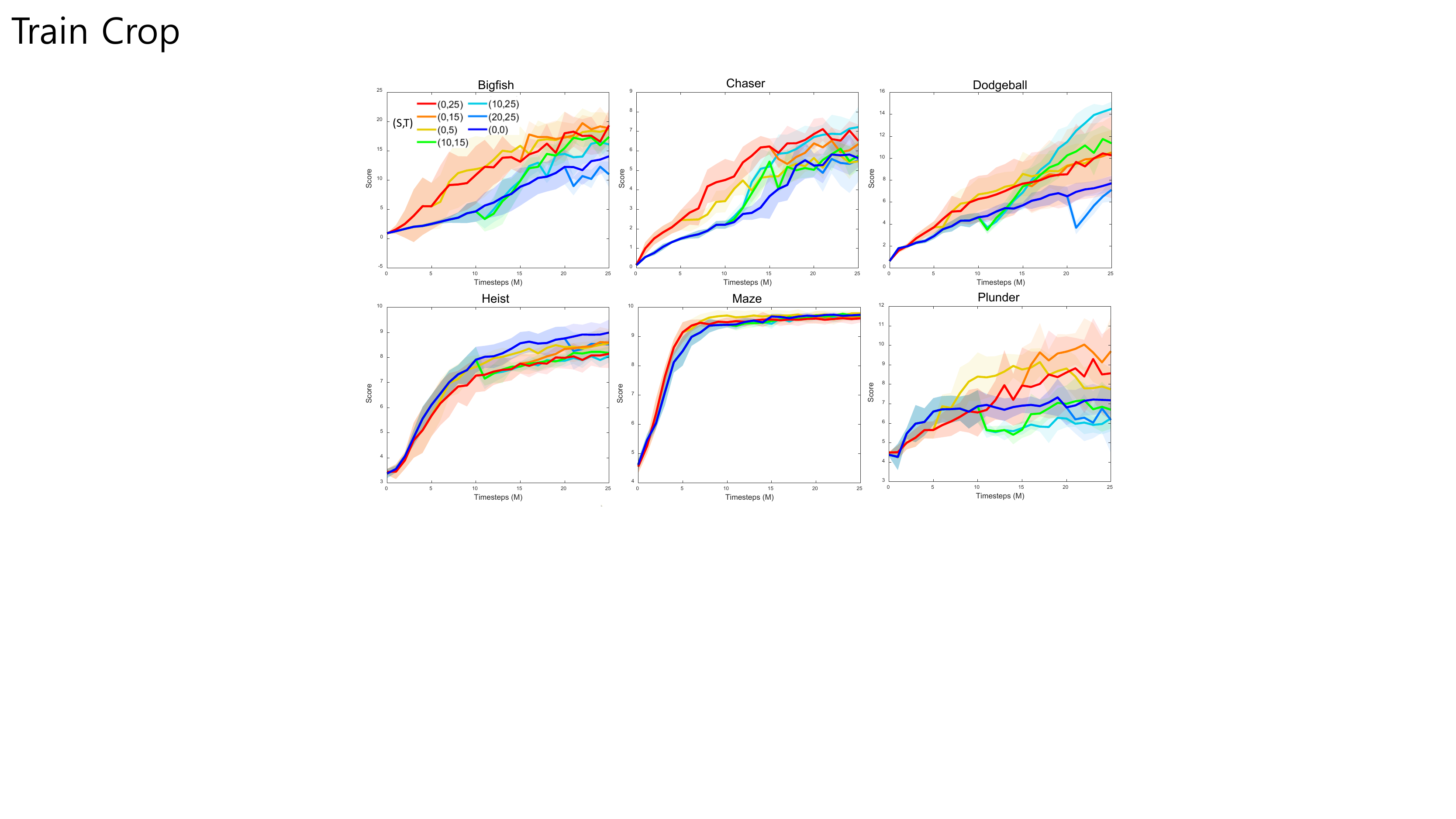}
    \caption{Comparison of training performance according to usage period of augmentation with \inda{} ({\it crop}): {\it Crop} improve the training performance in Bigfish, Chaser, Dodgeball, Plunder. Furthermore, interrupted augmentation is also improved similarly with (0, 25).}
    
\end{figure}
\begin{figure}[H]
    \centering
    \includegraphics[width=0.9\textwidth]{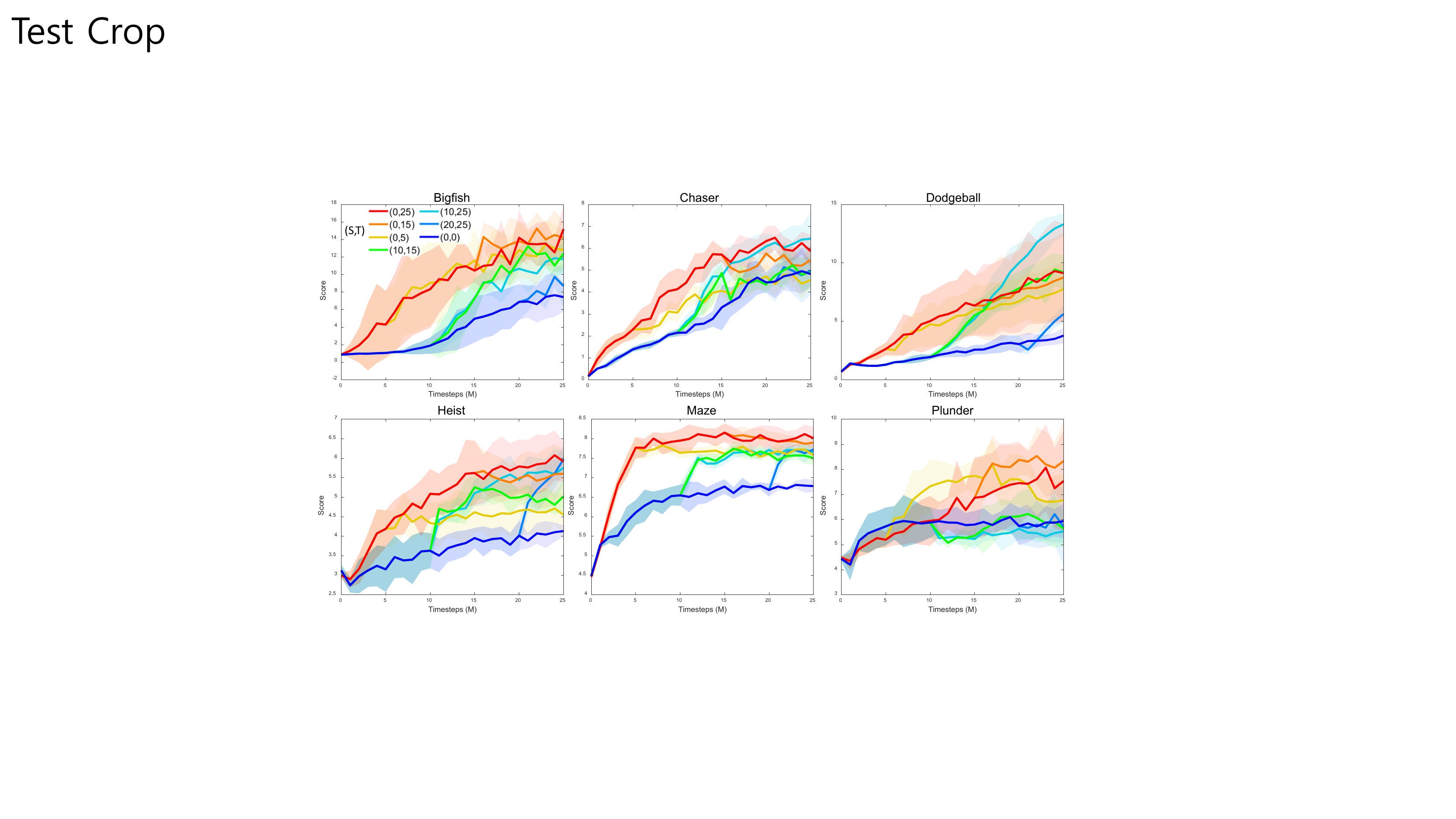}
    \caption{Comparison of generalization on unseen levels according to usage period of augmentation with \inda{} ({\it crop}): The generalization is improved by {\it crop}, and it is conserved after interrupted in Heist and Maze. Bigfish, Chaser, Dodgeball, and Plunder have similar curves with training. }
\end{figure}

\subsection{Color jitter}

\begin{figure}[H]
    \centering
    \includegraphics[width=0.84\textwidth]{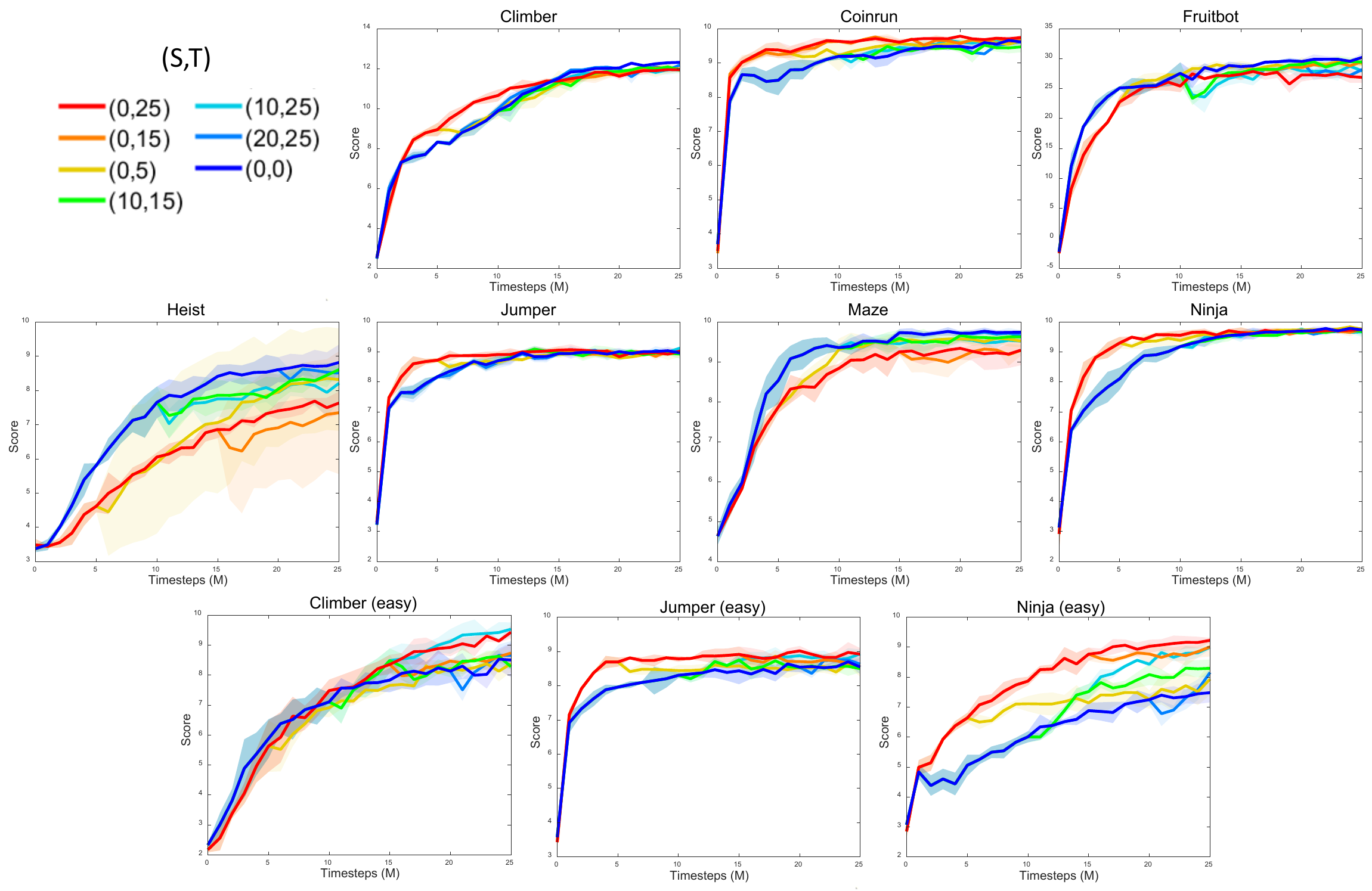}
    \caption{Comparison of training performance according to usage period of augmentation with \inda{} ({\it color jitter}): {\it Color jitter} does not impede the training as much as {\it random convolution} in most environments. However, {\it color jitter} helps the training in {\it easy} mode. }
\end{figure}

\begin{figure}[H]
    \centering
    \includegraphics[width=0.84\textwidth]{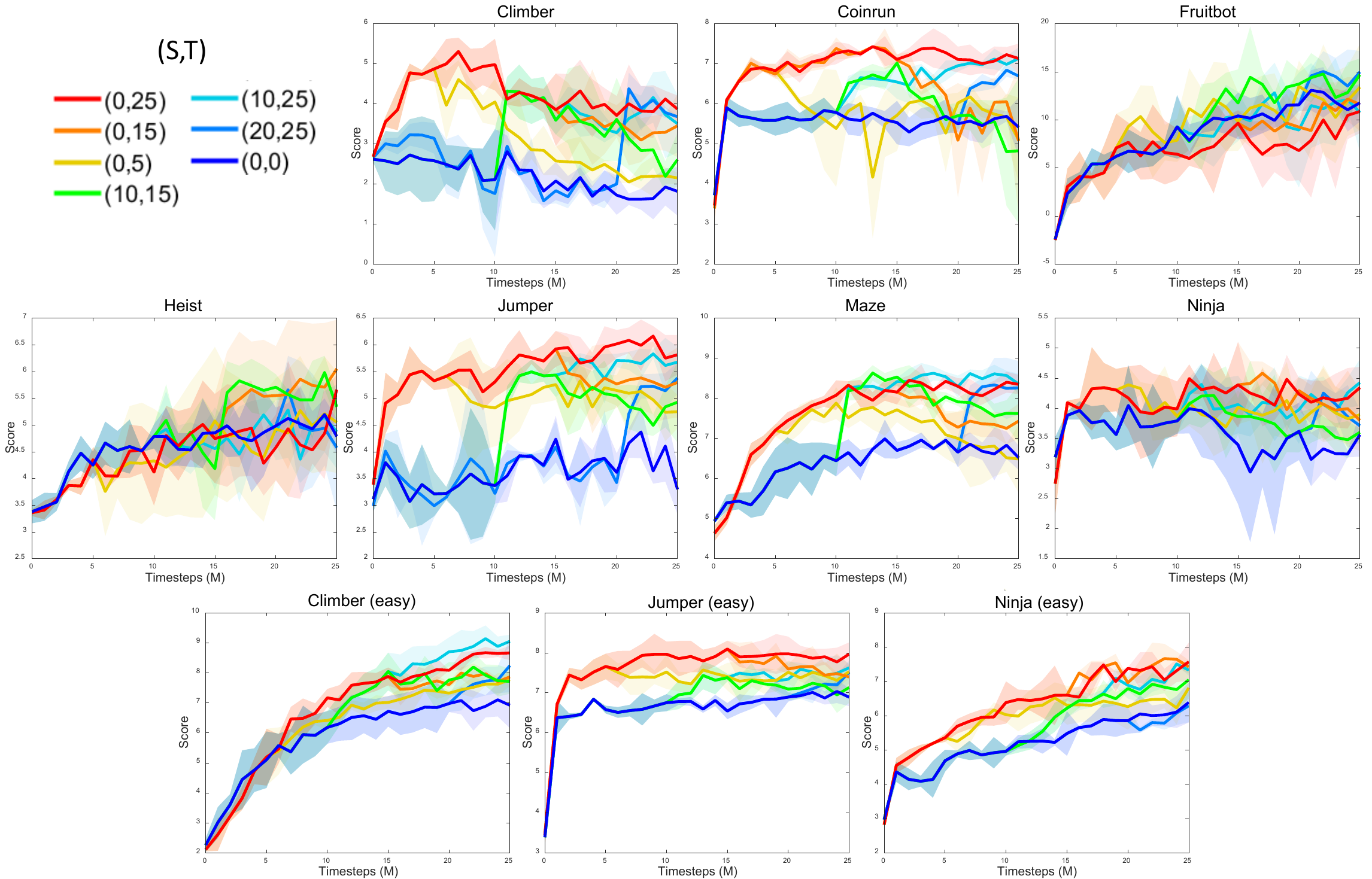}
    \caption{Comparison of generalization on unseen backgrounds according to usage period of augmentation with \inda{} ({\it color jitter}): Test performance is influenced by {\it color jitter} as the trend, which is similar to {\it random convolution}. }
    
\end{figure}

\section{Benchmark on Modified Open AI Procgen}
\label{sec:benchmark}
\begin{figure}[H]
    \centering
    \centering
        \includegraphics[width=0.45\textwidth]{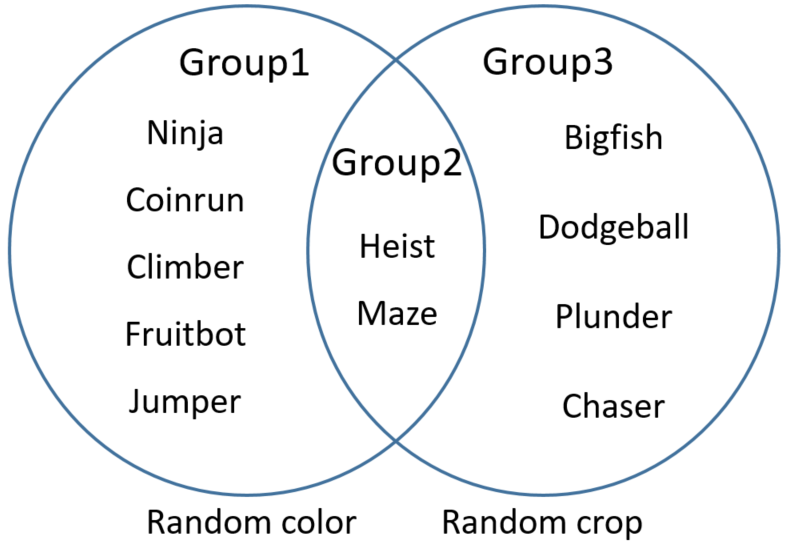}
        \caption{Group of OpenAI Procgen environments: We classify environments according to characteristics about a type of observation window and the importance of color. For Group1, we do not use random crop because they have agent centered window, thus the {\it random crop} can make hard to find agent. For Group3, we do not use random color, because they have important meaning in the color of objects, thus the transformation of color would make hard to learn information in color. Group2 can apply both of the transformations. }
        \label{fig:group}
\end{figure}

We compare the training and test performance on various environments with each augmentation. We also use DrAC \cite{raileanu2020automatic}, RAD \cite{laskin2020reinforcement}, DrAC+PAGrad as baselines. In every result, we train the agent for 25M timesteps, except the \exda{}. 
\exda{} is trained with 0.5M after training 20M with PPO. We also compare the average score after normalized by PPO's score and indicate the best score as bold. Mean and standard deviation is calculated after five runs. We show the result about {\it random conv, color jitter, random crop} in Table~\ref{table:Performance Benchmark All}. Additionally, we evaluate benchmark with {\it gray} and {\it cutout color} in Table~\ref{table: Benchmark-gray-cc}. For benchmark, we classify Procgen environments with each characteristic as Figure~\ref{fig:group}. Furthermore, we attach detail results on each environments with Oracle and Rand-FM \cite{lee2019network}.  Red one is the Oracle score, which is trained on test environments such as {\it easybg-test, easy-test}. For your information, RAD does not work well when using crop, because we use \cite{raileanu2020automatic}'s crop method which is different with \cite{laskin2020reinforcement}.

\begin{table}[ht]
    \centering
    \renewcommand{\multirowsetup}{\centering}
      \begin{tabular}{ cc|c|ccc|cc}
        \toprule
        Augmentation & Task & PPO  & RAD  & DrAC  & DrAC+PAGrad & \inda{} &  \exda{}\\
        \midrule
        \multirow{3}{20mm}{Grayscale} & Train & \bf{1.00} & 0.94 & 0.93  & 0.94  & 0.95 & 0.99  \\
                                 & Test-bg  & 1.00 & 1.04 & 1.03   & 1.04 & 0.97 & \bf{1.13} \\
         & Test-lv  & \bf{1.00} & 0.81 & 0.84 & 0.84 & 0.84 & 0.86 \\
        \midrule
        \multirow{3}{20mm}{Cutout color} & Train & \bf{1.00} & 0.72 & 0.82  & 0.84 & 0.76 & 0.94  \\  & Test-bg  & 1.00 & 1.33 & 1.27  & 1.29 & 1.19 & \bf{1.53} \\
        & Test-lv  & \bf{1.00} & 0.69 & 0.83  & 0.83 & 0.69 & 0.93 \\
       
        \bottomrule
      \end{tabular}
    \vspace*{3mm}
    \caption{Train and test score of \inda{} and \exda{} on Open AI Procgen, compared to baselines PPO, Drac \cite{raileanu2020automatic}, RAD \cite{laskin2020reinforcement}, DrAC+PAGrad. {\bf Boldface} indicates the best method.  }
    \label{table: Benchmark-gray-cc} 
\end{table}

\subsection{Random convolution}

\begin{table}[H]
\centering
\begin{tabular}{ccccc|cc}
\toprule
\multicolumn{1}{c}{Easy} & PPO   & DrAC  & Rand\_FM & RAD   & InDA  & ExDA  \\ \midrule
\multirow{2}{*}{Climber}  & \textbf{8.63}  & 8.33  & 8.27     & 7.93  & 8.5   & 8.1   \\
                          & \std0.462 & \std0.407 & \std0.187    & \std0.37  & \std0.291 & \std0.268 \\
\multirow{2}{*}{Jumper}   & 8.55  & 8.62  & 8.47     & 8.51  & \textbf{8.94}  & 8.5   \\
                          & \std0.168 & \std0.075 &\std0.13     & \std0.102 & \std0.09  & \std0.183 \\
\multirow{2}{*}{Ninja}    & 7.49  & 8.57  & 7.69     & 7.9   & \textbf{8.88}  & 7.03  \\
                          & \std0.421 & \std0.069 & \std0.529    & \std0.652 & \std0.343 & \std0.058 \\ \midrule
Avg                       & 1.00     & 1.04  & 0.99     & 0.99  & 1.07  & 0.96 \\ \bottomrule
\end{tabular}
\caption{Training performance benchmark on {\it easy} with {\it random convolution}. }
\end{table}

\begin{table}[H]
\centering
\begin{tabular}{ccccc|cc}
\toprule
\multicolumn{1}{c}{Easy} & PPO   & DrAC  & Rand\_FM & RAD   & InDA  & ExDA  \\ \midrule
\multirow{2}{*}{Climber} & 6.96  & 7.21  & 6.63     & 6.08  & \textbf{7.28}  & 7.06  \\
                         & \std0.651 & \std0.447 & \std0.39     & \std0.264 & \std0.341 & \std0.541 \\
\multirow{2}{*}{Jumper}  & 6.85  & \textbf{7.97}  & 6.7      & 6.74  & 7.94  & 7.54  \\
                         & \std0.192 & \std0.128 & \std0.167    & \std0.299 & \std0.185 & \std0.158 \\
\multirow{2}{*}{Ninja}   & 6.29  & 6.18  & 6.22     & 6.22  & \textbf{6.5}   & 5.56  \\
                         & \std0.529 & \std0.193 & \std0.57     & \std0.324 & \std0.191 & \std0.158 \\ \midrule
Avg                      & 1.00     & 1.06  & 0.97     & 0.95  & \textbf{1.08}  & 1    \\ \bottomrule
\end{tabular}
\caption{Test performance benchmark on unseen backgrounds ({\it easy, random convolution}). }
\end{table}

\begin{table}[H]
\centering
\begin{tabular}{cccccc|cc}
\toprule
          Easybg                & PPO   & Oracle & DrAC  & Rand-FM & RAD   & InDA  & ExDA  \\ \midrule
\multirow{2}{*}{Climber}  & \textbf{12.35} & 9.78  & 11.23 & 12.2    & 12.15 & 10.89 & 12.07 \\
                          & \std0.083 & \std0.306 & \std0.353 & \std0.128   & \std0.09  & \std0.162 & \std0.073 \\
\multirow{2}{*}{Coinrun}  & \textbf{9.64}  & 7.11  & 9.17  & 9.57    & 9.56  & 8.81  & 9.44  \\
                          & \std0.07  & \std0.205 & \std0.161 & \std0.126   & \std0.107 & \std0.992 & \std0.149 \\
\multirow{2}{*}{Fruitbot} & 29.78 & 29.74 & 26.07 & \textbf{30.19}   & 29.92 & 26.17 & 28.76 \\
                          & \std0.899 & \std0.443 & \std0.658 & \std0.512   & \std0.623 & \std0.575 & \std0.79  \\
\multirow{2}{*}{Heist}    & \textbf{9}     & 7.21  & 5.95  & 7.7     & 7.94  & 5.15  & 8.72  \\
                          & \std0.513 & \std0.27  & \std0.343 & \std0.6     & \std0.919 & \std0.614 & \std0.533 \\
\multirow{2}{*}{Jumper}   & 8.95  & 8.72  & 8.86  & 8.91    & \textbf{9.04}  & 8.78  & 8.94  \\
                          & \std0.066 & \std0.119 & \std0.088 & \std0.13    & \std0.135 & \std0.172 & \std0.048 \\
\multirow{2}{*}{Maze}     & \textbf{9.75}  & 8.56  & 8.1   & 9.61    & 9.51  & 9.12  & 9.73  \\
                          & \std0.513 & \std0.27  & \std0.343 & \std0.6     & \std0.919 & \std0.614 & \std0.533 \\
\multirow{2}{*}{Ninja}    & 9.75  & 7.81  & 9.43  & 9.75    & \textbf{9.78}  & 9.53  & 9.7   \\
                          & \std0.073 & \std0.422 & \std0.109 & \std0.084   & \std0.03  & \std0.113 & \std0.062 \\ \midrule
Avg                       & \textbf{1.00}     & 0.85  & 0.98  & 0.98    & 0.88  & 0.88  & 0.98  \\ \bottomrule
\end{tabular}
\caption{Training performance benchmark on {\it easybg} with  {\it random convolution}. }
\end{table}

\begin{table}[H]
\centering
\begin{tabular}{cccccc|cc}
\toprule
            Easybg              & PPO   & Oracle & DrAC  & Rand-FM & RAD   & InDA  & ExDA  \\ \midrule
\multirow{2}{*}{Climber}  & 1.97  & \textcolor{red}{9.78}  & 7.13  & 2       & 2.34  & 7.36  & \textbf{8.11}  \\
                          & \std0.51  & \std0.306 & \std0.419 & \std0.59    & \std1.258 & \std0.273 & \std0.457 \\
\multirow{2}{*}{Coinrun}  & 5.48  & \textcolor{red}{7.11}  & 7.54  & 5.65    & 5.48  & 7.14  & \textbf{7.81}  \\
                          & \std0.583 & \std0.205 & \std0.188 & \std0.216   & \std0.542 & \std0.479 & \std0.388 \\
\multirow{2}{*}{Fruitbot} & 10.83 & \textcolor{red}{29.74} & 19.77 & 15.19   & 11.61 & 21.93 & \textbf{23.57} \\
                          & \std1.908 & \std0.443 & \std0.77  & \std3.363   & \std4.615 & \std0.664 & \std0.745 \\
\multirow{2}{*}{Heist}    & 5.18  & \textcolor{red}{7.21}  & 5.47  & 5.03    & 4.78  & 4.96  & \textbf{8.15}  \\
                          & \std0.838 & \std0.27  & \std0.326 & \std0.6     & \std0.785 & \std0.777 & \std0.633 \\
\multirow{2}{*}{Jumper}   & 3.38  & \textcolor{red}{8.72}  & 8.14  & 4.12    & 3.77  & \textbf{8.16}  & 7.87  \\
                          & \std0.368 & \std0.119 & \std0.17  & \std0.514   & \std0.435 & \std0.231 & \std0.485 \\
\multirow{2}{*}{Maze}     & 6.48  & \textcolor{red}{8.56}  & 6.4   & 6.6     & 6.29  & 8.41  & \textbf{8.92}  \\
                          & \std0.523 & \std0.665 & \std0.419 & \std0.494   & \std0.466 & \std0.436 & \std0.155 \\
\multirow{2}{*}{Ninja}    & 3.83  & \textcolor{red}{7.81}  & 6.8   & 3.36    & 3.98  & 6.61  & \textbf{6.85}  \\
                          & \std0.462 & \std0.422 & \std0.243 & \std0.505   & \std0.44  & \std0.327 & \std0.25  \\ \midrule
Avg                       & 1.00     & \textcolor{red}{2.33}  & 1.86  & 1.08    & 1.04  & 1.92  & \textbf{2.11}  \\ \bottomrule
\end{tabular}
\caption{Test performance benchmark on unseen backgrounds ({\it easybg, random convolution}).}
\end{table}

\subsection{Crop}
\label{sec:crop-bench}
\begin{table}[H]
\centering
\begin{tabular}{cccc|cc}
\toprule
\multicolumn{1}{c}{Easybg}  & PPO           & DrAC  & RAD   & InDA           & ExDA  \\ \midrule
\multirow{2}{*}{Bigfish}   & 14.08         & 15.92 & 5.05  & \textbf{19.35} & 11.07 \\
                           & \std2.229         & \std1.535 & \std3.718 & \std2.792          & \std3.683 \\
\multirow{2}{*}{Chaser}    & 5.63          & 3.97  & 1.24  & \textbf{6.52}  & 4.81  \\
                           & \std0.467         & \std0.642 & \std0.253 & \std0.825          & \std0.325 \\
\multirow{2}{*}{Dodgeball} & 7.71          & 10.74 & 1.23  & \textbf{12.74} & 6.74  \\
                           & \std0.678         & \std0.711 & \std0.944 & \std1.729          & \std0.815 \\
\multirow{2}{*}{Heist}     & \textbf{9}    & 7.58  & 4.53  & 8.15           & 8.79  \\
                           & \std0.513         & \std0.11  & \std0.266 & \std0.57           & \std0.424 \\
\multirow{2}{*}{Maze}      & \textbf{9.75} & 9.03  & 3.95  & 9.63           & 9.72  \\
                           & \std0.033         & \std0.348 & \std3.418 & \std0.143          & \std0.026 \\
\multirow{2}{*}{Plunder}   & 7.18          & 10.73 & 0     & \textbf{10.29} & 6.59  \\
                           & \std0.73          & \std1     & \std0     & \std0.285          & \std1.108 \\ \midrule
\multicolumn{1}{c}{Avg}    & 1.00             & 1.08  & 0.28  & \textbf{1.25}  & 0.91  \\ \bottomrule
\end{tabular}
\caption{Training performance benchmark on {\it easybg} with {\it crop.} }
\end{table}

\begin{table}[H]
\centering
\begin{tabular}{cccc|cc}
\toprule
Easybg                       & PPO           & DrAC          & RAD   & InDA           & ExDA          \\ \midrule
\multirow{2}{*}{Bigfish}   & 7.43          & 13.63         & 4.93  & \textbf{15.19} & 6.35          \\
                           & \std1.65          & \std1.504         & \std3.696 & \std2.724          & \std2.466         \\
\multirow{2}{*}{Chaser}    & 4.83          & 3.59          & 1.2   & \textbf{5.86}  & 4.48          \\
                           & \std0.56          & \std0.519         & \std0.259 & \std0.745          & \std0.379         \\
\multirow{2}{*}{Dodgeball} & 3.78          & 9.26          & 1.11  & \textbf{11.92} & 3.79          \\
                           & \std0.659         & \std0.685         & \std0.831 & \std1.556          & \std0.748         \\
\multirow{2}{*}{Heist}     & \textbf{4.13} & 5.4           & 3.81  & \textbf{5.91}  & 5.35          \\
                           & \std0.146         & \std0.448         & \std0.412 & \std0.516          & \std0.22          \\
\multirow{2}{*}{Maze}      & \textbf{6.79} & 7.77          & 3.9   & \textbf{8.01}  & \textbf{7.74} \\
                           & \std0.158         & \std0.328         & \std3.377 & \std0.288          & \std0.054         \\
\multirow{2}{*}{Plunder}   & 5.94          & \textbf{9.49} & 0     & \textbf{8.98}  & 5.98          \\
                           & \std0.698         & \std0.605         & \std0     & \std0.369          & \std0.944         \\ \midrule
Avg                        & 1.00             & 1.519         & 0.459 & \textbf{1.798} & 1.094        \\ \bottomrule
\end{tabular}
\caption{Test performance benchmark on unseen levels ({\it easybg, crop}). }
\end{table}
\subsection{Color jitter}

\begin{table}[H]
\centering
\begin{tabular}{cccc|cc}
\toprule
              Easy           & PPO   & DrAC          & RAD   & InDA          & ExDA  \\ \midrule
\multirow{2}{*}{Climber} & 8.5   & \textbf{9.33} & 8.64  & 9.43          & 8.18  \\
                         & \std0.575 & \std0.212         & \std0.156 & \std0.21          & \std0.45  \\
\multirow{2}{*}{Jumper}  & 8.54  & 8.64          & 8.63  & \textbf{8.92} & 8.44  \\
                         & \std0.22  & \std0.135         & \std0.17  & \std0.174         & \std0.185 \\
\multirow{2}{*}{Ninja}   & 7.48  & 8.69          & 8.24  & \textbf{9.23} & 7.37  \\
                         & \std0.324 & \std0.331         & \std0.251 & \std0.081         & \std0.212 \\ \midrule
Avg                      & 1.00     & 1.09          & 1.04  & \textbf{1.13} & 0.98  \\ \bottomrule
\end{tabular}
\caption{Training performance benchmark on {\it easy} with {\it color jitter}. }
\end{table}

\begin{table}[H]
\centering
\begin{tabular}{cccc|cc}
\toprule
                Easy         & PPO   & DrAC  & RAD   & InDA          & ExDA  \\ \midrule
\multirow{2}{*}{Climber} & 6.92  & 8.53  & 8.37  & \textbf{8.66} & 8.14  \\
                         & \std0.761 & \std0.422 & \std0.023 & \std0.24          & \std0.477 \\
\multirow{2}{*}{Jumper}  & 6.89  & 7.58  & 7.86  & \textbf{7.97} & 7.25  \\
                         & \std0.223 & \std0.053 & \std0.297 & \std0.292         & \std0.131 \\
\multirow{2}{*}{Ninja}   & 6.39  & 6.79  & 7.31  & \textbf{7.57} & 6.2   \\
                         & \std0.585 & \std0.32  & \std0.613 & \std0.555         & \std0.085 \\ \midrule
Avg                      & 1.00     & 1.13  & 1.16  & \textbf{1.2}  & 1.07  \\ \bottomrule
\end{tabular}
\caption{Test performance benchmark on unseen backgrounds ({\it easy, color jitter}). }
\end{table}

\begin{table}[H]
\centering
\begin{tabular}{ccccc|cc}
\toprule
Easybg                     & PPO            & Oracle & DrAC           & RAD   & InDA          & ExDA          \\ \midrule
\multirow{2}{*}{Climber}  & \textbf{12.31} & 9.85   & 11.84          & 12    & 11.94         & 12.04         \\
                          & \std0.092          & \std0.298  & \std0.223          & \std0.256 & \std0.071         & \std0.152         \\
\multirow{2}{*}{Coinrun}  & 9.61           & 7.2    & 8.94           & 8.62  & \textbf{9.74} & 9.45          \\
                          & \std0.074          & \std0.195  & \std0.285          & \std0.091 & \std0.05          & \std0.09          \\
\multirow{2}{*}{Fruitbot} & 30.2           & 29.69  & \textbf{30.05} & 29.48 & 26.87         & 29            \\
                          & \std0.691          & \std0.619  & \std0.611          & \std0.507 & \std0.912         & \std0.878         \\
\multirow{2}{*}{Heist}    & \textbf{8.82}  & 7.33   & 7.22           & 6.89  & 7.63          & 8.53          \\
                          &\std0.523          & \std0.308  & \std0.76           & \std0.348 & \std0.338         & \std0.307         \\
\multirow{2}{*}{Jumper}   & 8.97           & 8.67   & 8.9            & 8.94  & \textbf{9.03} & \textbf{9.03} \\
                          & \std0.075          & \std0.132  & \std0.05           & \std0.029 & \std0.123         & \std0.086         \\
\multirow{2}{*}{Maze}     & \textbf{9.75}  & 8.08   & 9.46           & 9.46  & 9.3           & 9.67          \\
                          & \std0.035          & \std0.101  & \std0.404          & \std0.184 & \std0.379         & \std0.111         \\
\multirow{2}{*}{Ninja}    & 9.74           & 7.56   & 9.52           & 9.65  & \textbf{9.75} & 9.54          \\
                          & \std0.087          & \std0.286  & \std0.393          & \std0.112 & \std0.046         & \std0.171         \\ \midrule
Avg                       & \textbf{1.00}     & 0.85   & 0.95           & 0.94  & 0.96          & 0.98         \\ \bottomrule
\end{tabular}
\caption{Training performance benchmark on {\it easbg} with {\it color jitter}. }
\end{table}

\begin{table}[H]
\centering
\begin{tabular}{ccccc|cc}
\toprule
Easybg                      & PPO           & Oracle & DrAC          & RAD   & InDA          & ExDA          \\ \midrule
\multirow{2}{*}{Climber}  & 1.82 & \textcolor{red}{9.85}   & \textbf{5.05} & 4.31  & 4.25          & 4.34          \\
                          & \std0.605         & \std0.298  & \std0.407         & \std0.492 & \std0.338         & \std0.856         \\
\multirow{2}{*}{Coinrun}  & 5.42          & \textcolor{red}{7.2}    & 6.46          & 6.47  & \textbf{7.13} & 6.53          \\
                          & \std0.744         & \std0.195  & 0\std.526         & \std0.194 & \std0.372         & \std0.375         \\
\multirow{2}{*}{Fruitbot} & 11.78         & \textcolor{red}{29.69}  & 9.49 & 8.51  & 10.88         & \textbf{18}   \\
                          & \std1.949         & \std0.619  & \std8.098         & \std1.941 & \std2.263         & \std7.442         \\
\multirow{2}{*}{Heist}    & 4.79 & \textcolor{red}{7.33}    & 5.65         & 5.39  & \textbf{5.66} & 5.43          \\
                          & \std0.323         & \std0.308  & \std0.984         & \std0.745 & \std0.271         & \std0.508         \\
\multirow{2}{*}{Jumper}   & 3.3           & \textcolor{red}{8.6}   & 5.65          & 5.67  & \textbf{5.81} & 5.31 \\
                          & \std0.467         & \std0.132  & \std0.09          & \std0.953 & \std0.369         & \std0.351         \\
\multirow{2}{*}{Maze}     & 6.52 &  \textcolor{red}{8.08}     &8.22        & 8.26  & 8.35          & \textbf{8.65} \\
                          & \std0.304         & \std0.101  & \std0.455         & \std0.175 & \std0.238         & \std0.017         \\
\multirow{2}{*}{Ninja}    & 3.56          & \textcolor{red}{7.56}   & 4.22          & 4.18  & \textbf{4.34} & 4.07          \\
                          & \std0.363         & \std0.286  & \std0.487         & \std0.475 & \std0.345         & \std0.332         \\ \midrule
Avg                       & 1.00    & \textcolor{red}{2.4}   & 1.44          & 1.37  & 1.43          & \textbf{1.48 }         \\ \bottomrule
\end{tabular}
\caption{Test performance benchmark on unseen backgrounds ({\it easybg, color jitter}). }
\end{table}

\subsection{Gray}

\begin{table}[ht]
\centering
\begin{tabular}{ccccc|cc}
\toprule
Easybg                    & PPO            & Oracle   & DrAC  & RAD& InDA     & ExDA          \\ \midrule
\multirow{2}{*}{Climber}  & \textbf{12.31} & 9.85             & 11.12 & 11.84 & 11.9 & 12.06         \\
                          & \std0.092          & \std0.298           & \std0.26  & \std0.505  & \std0.115& \std0.03          \\
\multirow{2}{*}{Coinrun}  & 9.61           & 7.2     & 9.53  & 9.49 & \textbf{9.74} & 9.48          \\
                          & \std0.074          & \std0.195          & \std0.135 & \std0.188 & \std0.046  & \std0.08          \\
\multirow{2}{*}{Fruitbot} & \textbf{30.2}  & 29.69           & 30.01 & 29.6  & 28.03 & 29.32         \\
                          & \std0.691          & \std0.619          & \std0.572 & \std0.27 & \std0.994  & \std0.937         \\
\multirow{2}{*}{Heist}    & \textbf{8.82}  & 7.33            & 6.24  & 6.53  & 5.51  & 8.51          \\
                          & \std0.523          & \std0.308           & \std0.214 & \std0.474 & \std0.146 & \std0.225         \\
\multirow{2}{*}{Jumper}   & 8.54           & 8.67    & 8.91  & 8.93 & \textbf{9.18}  & 8.95          \\
                          & \std0.22           & \std0.132            & \std0.19  & \std.247 & \std0.18 & \std0.075         \\
\multirow{2}{*}{Maze}     & \textbf{9.75}  & 8.08             & 9.46  & 9.48 & 9.2  & \textbf{9.75} \\
                          & \std0.035          & \std0.101          & \std0.192 & \std0.08 & \std0.367   & \std0.087         \\
\multirow{2}{*}{Ninja}    & \textbf{9.74}  & 7.56              & 9.73  & 9.61 & 9.6 & 9.72          \\
                          & \std0.087          & \std0.286           & \std0.045 & \std0.096 & \std0.081 & \std0.021         \\ \midrule
Avg                       & \textbf{1}     & 0.85   & 0.93          & 0.94  & 0.95  & 0.99          \\ \bottomrule
\end{tabular}
\caption{Training performance benchmark on {\it easybg} with {\it gray}. }
\end{table}

\begin{table}[H]
\centering
\begin{tabular}{ccccc|cc}
\toprule
Easybg                    & PPO           & Oracle           & RAD            & DrAC & InDA  & ExDA          \\ \midrule
\multirow{2}{*}{Climber}  & 1.82          & \textcolor{red}{9.85}            & 1.75           & 1.81 & 1.24  & \textbf{2.45} \\
                          & \std0.605         &\std 0.298           & \std0.654          & \std0.211 & \std0.502 & \std0.727         \\
\multirow{2}{*}{Coinrun}  & 5.42          & \textcolor{red}{7.2 }   & 5.34           & 5.31 & \textbf{6.05}  & 5.79          \\
                          & \std0.744         & \std0.195           & \std0.751          & \std0.501 & \std0.465 & \std0.061         \\
\multirow{2}{*}{Fruitbot} & 11.78         & \textcolor{red}{29.69}           & \textbf{17.57} & 15.47 & 15.12 & 15.81         \\
                          & \std1.949         & \std0.619           & \std0.191          & \std1.449 & \std0.958 & \std0.11          \\
\multirow{2}{*}{Heist}    & 4.79          & \textcolor{red}{7.33}            & \textbf{5.43}  & 5.15 & 4.32  & 5.1           \\
                          & \std0.323         & \std0.308          & \std0.18           & \std0.172 & \std0.112  & \std0.504         \\
\multirow{2}{*}{Jumper}   & \textbf{6.89} & \textcolor{red}{8.67}            & 2.7            & 4.07 & 3.55  & 4.47          \\
                          & \std0.223         & \std0.132          & \std0.894          & \std0.46  & \std0.992  & \std0.415         \\
\multirow{2}{*}{Maze}     & 6.52          & \textcolor{red}{8.08}            & 7.77           & 7.93 & 7.67  & \textbf{8.33} \\
                          & \std0.304         & \std0.101          & \std0.611          & \std0.104 & \std0.312  & \std0.119         \\
\multirow{2}{*}{Ninja}    & 3.56          & \textcolor{red}{7.56}            & 3.72           & 3.91 & 4.02  & \textbf{4.03} \\
                          & \std0.363         & \std0.286           & \std0.131          & \std0.62 & \std0.666  & \std0.071         \\ \midrule
Avg                       & 1             & \textcolor{red}{2.2 }           & 1.03           & 1.04  & 0.97  & \textbf{1.13} \\ \bottomrule
\end{tabular}
\caption{Test performance benchmark on unseen backgrounds ({\it easybg, gray}). }
\end{table}

\begin{table}[H]
\centering
\begin{tabular}{cccc|cc}
\toprule
Easy                     & PPO          & DrAC     & RAD      & InDA             & ExDA     \\ \midrule
\multirow{2}{*}{Climber} & \textbf{8.5} & 6.95     & 7.55     & 7.22             & 8.05     \\
                         & \std0.575        &\std 0.547    & \std0.256    & \std0.312            & \std0.461    \\
\multirow{2}{*}{Jumper}  & 8.54         & 8.4      & 8.58     & \textbf{8.85}    & 8.5      \\
                         & \std0.22         & \std0.224    & \std0.199    & \std0.015            & \std0.224    \\
\multirow{2}{*}{Ninja}   & 7.48         & 6.67     & 7.1      & \textbf{8.91}    & 7.05     \\
                         & \std0.324        & \std0.435    & \std0.718    & \std0.165            & \std0.24     \\ \midrule
Avg                      & 1            & 0.9 & 0.95 & \textbf{1.026} & 0.96 \\ \bottomrule

\end{tabular}
\caption{Training performance benchmark on {\it easybg} with {\it gray}. }
\end{table}

\begin{table}[H]
\centering
\begin{tabular}{cccc|cc}
\toprule
Easy                     & PPO           & DrAC     & RAD      & InDA          & ExDA          \\ \midrule
\multirow{2}{*}{Climber} & 6.92          & 4.49     & 5.57     & 5.11          & \textbf{7.24} \\
                         & \std0.761         & \std0.332    & \std0.307    & \std0.483         & \std0.721         \\
\multirow{2}{*}{Jumper}  & \textbf{6.89} & 5.38     & 6.59     & 6.35          & 6.87          \\
                         & \std0.223         & \std0.215    & \std0.055    & \std0.234         & \std0.182         \\
\multirow{2}{*}{Ninja}   & 6.39          & 5.67     & 5.14     & \textbf{6.84} & 6.01          \\
                         & \std0.585         & \std0.318    & \std0.628    & \std0.206         & \std0.651         \\\midrule
Avg                      & \textbf{1}    & 0.77 & 0.86 & 0.91      & 0.99     \\ \bottomrule
\end{tabular}
\caption{Test performance benchmark on unseen backgrounds ({\it easy, gray}). }
\end{table}

\subsection{Cutout color}
\begin{table}[H]
\centering
\begin{tabular}{ccccc|cc}
\toprule
Easybg                     & PPO           & Oracle & DrAC  & RAD   & InDA         & ExDA  \\ \midrule
\multirow{2}{*}{Climber}  & \textbf{12.31} & 9.85   & 11.92 & 8.26  & 11.76        & 12.07 \\
                          & \std0.092          & \std0.298  & \std0.158 & \std0.663 & \std0.027        & \std0.127 \\
\multirow{2}{*}{Coinrun}  & 9.61           & 7.2    & 9.23  & 8.07  & \textbf{9.7} & 9.39  \\
                          & \std0.074          & \std0.195  & \std0.323 & \std0.645 & \std0.084        & \std0.012 \\
\multirow{2}{*}{Fruitbot} & \textbf{30.2}  & 29.69  & 29.73 & 29.2  & 27.18        & 28.95 \\
                          & \std0.691          & \std0.619  & \std0.898 & \std0.64  & \std1.302        & \std0.907 \\
\multirow{2}{*}{Heist}    & \textbf{8.82}  & 7.33   & 8.47  & 6.25  & 6.1          & 8.65  \\
                          & \std0.523          & \std0.308  & \std0.397 & \std0.704 & \std0.693        & \std0.21  \\
\multirow{2}{*}{Jumper}   & 8.97           & 8.67   & 8.87  & 8.75  & \textbf{9.1} & 8.91  \\
                          & \std0.075          & \std0.132  & \std0.123 & \std0.131 & \std0.081        & \std0.053 \\
\multirow{2}{*}{Maze}     & \textbf{9.75}  & 8.08   & 9.41  & 9.17  & 9.27         & 9.74  \\
                          & \std0.035          & \std0.101  & \std0.134 & \std0.118 & \std0.125        & \std0.133 \\
\multirow{2}{*}{Ninja}    & \textbf{9.74}  & 7.56   & 9.65  & 7.17  & 9.72         & 9.7   \\
                          & \std0.087          & \std0.286  & \std0.138 & \std1.993 & \std0.02         & \std0.02  \\
\multirow{2}{*}{Bigfish}   & \textbf{13.89} & 13.22 & 2.54  & 5.19  & 1.95  & 11.22 \\
                           & \std3.127          & \std1.488 & \std0.13  & \std3.658 & \std0.311 & \std3.66  \\
\multirow{2}{*}{Chaser}    & \textbf{5.49}  & 3.04  & 2.88  & 1.98  & 3.34  & 5     \\
                           & \std0.562          & \std0.183 & \std0.699 & \std0.112 & \std0.755 & \std0.187 \\
\multirow{2}{*}{Dodgeball} & \textbf{7.76}  & 5.74  & 5.71  & 5.98  & 2.79  & 6.57  \\
                           & \std0.859          & \std1.118 & \std1.008 & \std0.103 & \std1.612 & \std0.693 \\
\multirow{2}{*}{Plunder}   & \textbf{7.15}  & 6.05  & 5.43  & 4.34  & 4.92  & 6.87  \\
                           & \std0.95           & \std0.58  & \std0.082 & \std0.24  & \std0.625 & \std1.255 \\ \midrule
Avg                        & \textbf{1.00}     & 0.82  & 0.82  & 0.72  & 0.76  & 0.94  \\ \bottomrule         
\end{tabular}
\caption{Training performance benchmark on {\it easybg} with {\it cutout color}. }
\end{table}

\begin{table}[H]
\centering
\begin{tabular}{ccccc|cc}
\toprule
Easy                    & PPO   & Oracle & DrAC  & RAD   & InDA          & ExDA  \\ \midrule
\multirow{2}{*}{Climber} & 8.5   & 9.85   & 7.69  & 6.67  & \textbf{9.02} & 8.02  \\
                         & \std0.575 & \std0.298  & \std0.237 & \std0.381 & \std0.473         & \std0.506 \\
\multirow{2}{*}{Jumper}  & 7.48  & 7.56   & 6.28  & 5.6   & \textbf{8.57} & 7.41  \\
                         & \std0.324 & \std0.286  & \std0.257 & \std0.276 & \std0.122         & \std0.125 \\
\multirow{2}{*}{Ninja}   & 8.54  & 8.67   & 8.45  & 8.32  & \textbf{8.93} & 8.53  \\
                         & \std0.22  & \std0.132  & \std0.183 & \std0.051 & \std0.166         & \std0.095 \\ \midrule
Avg                      & 1.00     & 1.06   & 0.91  & 0.84  & \textbf{1.08} & 0.98  \\ \bottomrule
\end{tabular}
\caption{Training performance benchmark on {\it easy} with {\it cutout color}. }
\end{table}

\begin{table}[H]
\centering
\begin{tabular}{ccccc|cc}
\toprule
Easybg                  & PPO           & Oracle & DrAC          & RAD            & InDA  & ExDA          \\ \midrule
\multirow{2}{*}{Climber}  & 1.82  & \textcolor{red}{9.85}   & 3.54          & 3.97           & 3.4   & \textbf{4.29} \\
                          & \std0.605 & \std0.298  & \std0.164         & \std0.999          & \std0.645 & \std0.154         \\
\multirow{2}{*}{Coinrun}  & 5.42  & \textcolor{red}{7.2}    & 5.87          & 5.93           & 6.2   & \textbf{6.41} \\
                          & \std0.744 & \std0.195  & \std.251         & \std0.061          & \std0.357 & \std0.131         \\
\multirow{2}{*}{Fruitbot} & 11.78 & \textcolor{red}{29.69}  & 18.18         & \textbf{19.24} & 17.69 & 17.7          \\
                          & \std1.949 & \std0.619  & \std3.744         & \std3.385          & \std4.026 & \std0.888         \\
\multirow{2}{*}{Heist}    & 4.79  & \textcolor{red}{7.33}   & 6.6           & 5.76           & 4.97  & \textbf{7.51} \\
                          & \std0.323 & \std0.308  & \std.092         & \std0.551          & \std0.33  & \std0.119         \\
\multirow{2}{*}{Jumper}   & 3.3   & \textcolor{red}{8.67}   & 4.99          & 5.48           & 5.43  & \textbf{6.02} \\
                          & \std0.467 & \std0.132  & \std0.114         & \std0.28           & \std1.116 & \std0.235         \\
\multirow{2}{*}{Maze}     & 6.52  & \textcolor{red}{8.08}   & 7.33          & 7.66           & 7.01  & \textbf{7.83} \\
                          & \std0.304 & \std0.101  & \std0.223         & \std.243          & \std0.17  & \std0.22          \\
\multirow{2}{*}{Ninja}    & 3.56  & \textcolor{red}{7.56}   & \textbf{4.29} & 3.96           & 3.75  & 3.76          \\
                          & \std0.363 & \std0.286  & \std0.245         & \std0.152          & \std0.333 & \std0.348         \\ 
\multirow{2}{*}{Bigfish}   & 3.4           & \textcolor{red}{13.22} & 1.29  & 2.5   & 1.29  & \textbf{4.49} \\
                           & \std0.487         & \std1.488 & \std0.08  & \std2.331 & \std0.152 & \std0.776         \\
\multirow{2}{*}{Chaser}    & 0.91          & \textcolor{red}{3.04 } & 1.08  & 1.13  & 1.68  & \textbf{1.73} \\
                           & \std0.061         & \std0.183 & \std0.038 & \std0.157 & \std0.305 & \std0.698         \\
\multirow{2}{*}{Dodgeball} & 2.17          & \textcolor{red}{5.74 } & 3.92  & 4.02  & 1.97  & \textbf{4.37} \\
                           & \std0.53          & \std1.118 & \std0.53  & \std0.345 & \std1.098 & \std0.527         \\
\multirow{2}{*}{Plunder}   & \textbf{6.87} & \textcolor{red}{6.05 } & 5.27  & 4.77  & 4.71  & 6.45          \\
                           & \std0.933         & \std0.58  & \std0.208 & \std0.612 & \std0.622 & \std1.232         \\ \midrule
Avg                        & 1.00             & \textcolor{red}{2.51}  & 1.27  & 1.33  & 1.19  & \textbf{1.53} \\ \bottomrule
\end{tabular}
\caption{Test performance benchmark on unseen backgrounds ({\it easybg, cutout color}). }
\end{table}

\begin{table}[H]
\centering
\begin{tabular}{ccccc|cc}
\toprule
Easy                  & PPO   & Oracle & DrAC  & RAD   & InDA  & ExDA  \\ \midrule
\multirow{2}{*}{Climber} & 6.92  & \textcolor{red}{9.85}   & 6.54  & 5.24  & \textbf{7.61} & 7.25          \\
                         & \std0.761 & \std0.298  & \std0.213 & \std0.417 & \std0.486         & \std0.325         \\
\multirow{2}{*}{Jumper}  & 6.39  & \textcolor{red}{7.56}   & 5.06  & 4.9   & \textbf{6.71} & 5.78          \\
                         & \std0.585 & \std0.286  & \std0.137 & \std0.382 & \std0.352         & \std0.488         \\
\multirow{2}{*}{Ninja}   & 6.89  & \textcolor{red}{8.67}   & 6.88  & 6.79  & 6.81          & \textbf{6.92} \\
                         & \std0.223 & \std0.132  & \std0.083 & \std0.278 & \std0.355         & \std0.212         \\ \midrule
Avg                      & 1.00     & \textcolor{red}{1.29}   & 0.91  & 0.84  & \textbf{1.05} & 0.99   \\ \bottomrule
\end{tabular}
\caption{Test performance benchmark on unseen backgrounds ({\it easy, cutout color}). }
\end{table}

\begin{table}[H]
\centering
\begin{tabular}{cccc|cc}
\toprule
Easybg                  & PPO            & DrAC          & RAD   & InDA  & ExDA  \\ \midrule
\multirow{2}{*}{Climber}  & \textbf{11.14} & 10.77         & 7.26  & 9.45  & 10.75 \\
                          &\std0.077          & \std0.279         & \std0.843 & \std0.193 & \std0.114 \\
\multirow{2}{*}{Coinrun}  & \textbf{8.64}  & 8.36          & 6.89  & 7.76  & 8.32  \\
                          & \std0.05           & \std0.348         & \std0.503 & \std0.096 & \std0.224 \\
\multirow{2}{*}{Fruitbot} & \textbf{28.26} & 26.88         & 26.22 & 23.79 & 26.33 \\
                          & \std0.461          & \std1.276         & \std1.258 & \std0.971 & \std0.894 \\
\multirow{2}{*}{Heist}    & \textbf{4.07}  & 3.92          & 2.27  & 2.15  & 3.93  \\
                          & \std0.07           & \std0.276         & \std0.448 & \std0.553 & \std0.184 \\
\multirow{2}{*}{Jumper}   & \textbf{7.38}  & 7.32          & 6.98  & 6.68  & 7.25  \\
                          & \std0.15           & \std0.195         & \std0.199 & \std0.24  & \std0.117 \\
\multirow{2}{*}{Maze}     & 6.8            & \textbf{6.84} & 6.04  & 5.91  & 6.17  \\
                          & \std0.2            & \std0.137         & \std0.258 & \std0.03  & \std0.162 \\
\multirow{2}{*}{Ninja}    & 8.56           & \textbf{8.63} & 6.28  & 7.81  & 8.34  \\
                          & \std0.061          & \std0.132         & \std1.866 & \std0.21  & \std0.119 \\ 
\multirow{2}{*}{Bigfish}   & \textbf{7.16} & 0.91  & 2.29  & 0.95  & 6.04  \\
                           & \std2.263         & \std0.037 & \std2.306 & \std0.06  & \std2.783 \\
\multirow{2}{*}{Chaser}    & \textbf{4.54} & 2.61  & 1.8   & 2.47  & 4.22  \\
                           & \std0.503         & \std0.509 & \std0.12  & \std0.506 & \std0.331 \\
\multirow{2}{*}{Dodgeball} & \textbf{3.78} & 2.71  & 2.53  & 1.26  & 2.82  \\
                           & \std0.823         & \std0.362 & \std0.135 & \std0.48  & \std0.593 \\
\multirow{2}{*}{Plunder}   & \textbf{5.99} & 5.08  & 4.07  & 4.55  & 5.83  \\
                           & \std0.814         & \std0.305 & \std0.479 & \std0.393 & \std1.061 \\ \midrule
Avg                        & \textbf{1.00}    & 0.83  & 0.69  & 0.69  & 0.93  \\ \bottomrule              
\end{tabular}
\caption{Test performance benchmark on unseen levels ({\it easybg, cutout color}).}
\end{table}

\begin{table}[H]
\centering
\begin{tabular}{cccc|cc}
\toprule
Easy                  & PPO   & DrAC          & RAD   & InDA  & ExDA          \\ \midrule
\multirow{2}{*}{Climber} & 5.45  & \textbf{5.9}  & 5.3   & 4.26  & 5.71          \\
                         & \std0.77  & \std0.352         & \std0.307 & \std0.122 & \std0.303         \\
\multirow{2}{*}{Jumper}  & 5.81  & \textbf{6.01} & 4.93  & 4.56  & 5.43          \\
                         & \std0.227 & \std0.389         & \std0.08  & \std0.161 & \std0.333         \\
\multirow{2}{*}{Ninja}   & 5.77  & 5.67          & 5.8   & 5.65  & \textbf{5.87} \\
                         & \std0.09  & \std0.023         & \std0.071 & \std0.166 & \std0.079         \\ \midrule
Avg                      & 1.00     & \textbf{1.03} & 0.94  & 0.85  & 1            \\ \bottomrule
\end{tabular}
\caption{Test performance benchmark on unseen levels ({\it easy, cutout color}). }
\end{table}

\section{Primitive evaluation on DeepMind Control Suite with SAC}

\begin{figure}[!t]
    \centering
    \centering
    \subfigure[]{
        \includegraphics[width=0.31\textwidth]{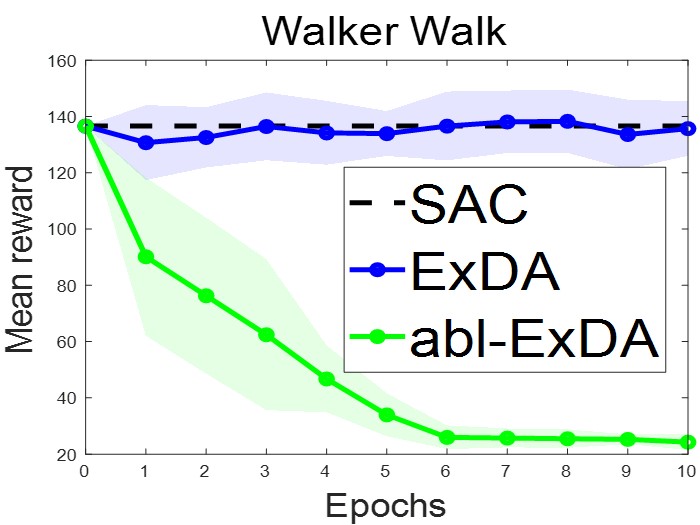}
        \label{fig:walkerwalk}
        }
    \subfigure[]{
        \includegraphics[width=0.31\textwidth]{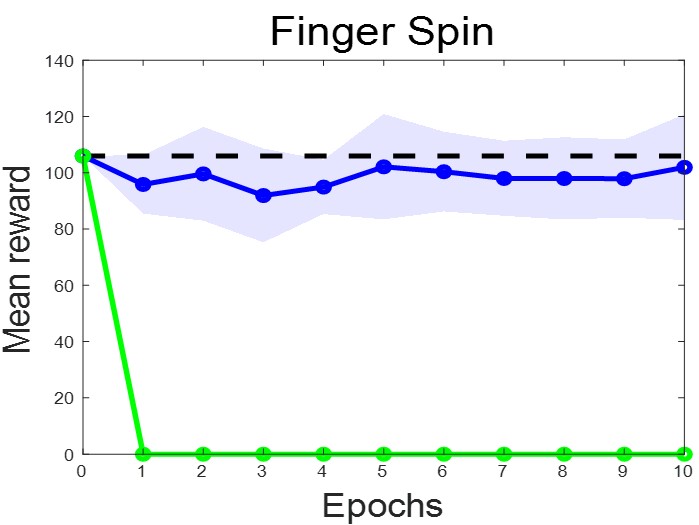}
        \label{fig:fingerspin}
        }
    \subfigure[]{
        \includegraphics[width=0.31\textwidth]{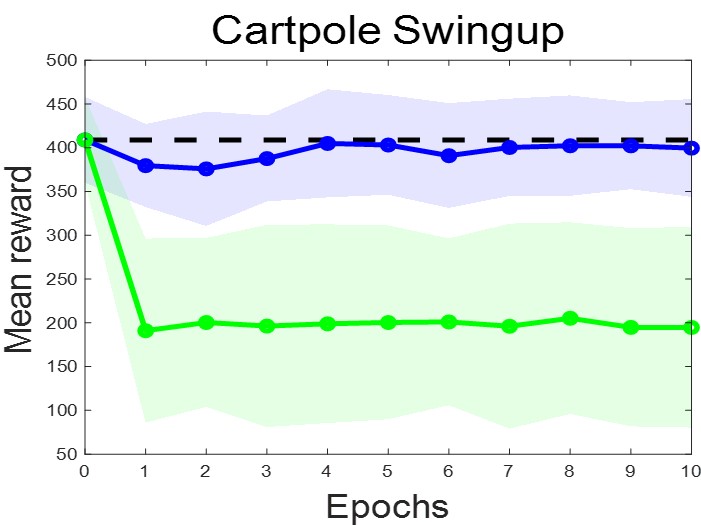}
        \label{fig:cartpoleswingup}
        }
    \caption{Ablation of \exda{} on DMcontrol }
    \label{fig:dmc}
\end{figure}

We experiment on DeepMind Control Suite (DMC) with a preliminary experiment which provides justification for our proposed method \exda{} due to the limited time and computation resources. We first note that \cite{raileanu2020automatic} has demonstrated that UCB-DrAC, which is similar to UCB-\inda{}, can accelerate RL training in DMC. Hence, in order to justify our proposed method UCB-\exda{}, which combines UCB-\inda{} and \exda{}, it would be sufficient to show the existence of the cases showing a benefit of \exda{} in DMC. Furthermore, we use SAC as the base RL algorithm instead of PPO to show the versatility of our method \exda{}. SAC is off-policy actor-critic RL algorithm which trains actor and critic networks separately with a replay buffer. Thus, we can use stored data in the replay buffer after RL training, and also we can only distill the output of the actor-network, because the actor-network determines the policy $\pi$ from observations, while the critic-network computes the value function $V$ from observations. We denote the actor-network parameters using $\theta$ same as the notation of PPO in Section~\ref{sec:method}. 
We ablate the effect of \exda{} to compare w/ and w/o inconsistency loss of policy on original observations. Thus \exda{} is trained with Eq~\ref{eq:L-DA} after SAC, but ablated \exda{} is trained with Eq~\ref{eq:ablated-exda}.  

\begin{align} \label{eq:ablated-exda}
L_{abl-\exda{}}(\theta, \phi; \theta_{\text{old}}) := 
L_{\text{dis}}(\theta, \phi; \theta_{\text{old}}) \;.
\end{align} 
Both methods are trained 10 epochs after 10K time-steps SAC training and detailed hyper-parameters are described below the table. 
We compare the mean reward of each method SAC, \exda{}, abl-\exda{} on three environments {\it Walker-Walk, Finger-Spin, Cartpole-Swingup} and 5 seeds. We evaluate each reward from 50 evaluation episodes in train environments. As below Figure~\ref{fig:dmc}, \exda{} almost maintains the mean reward of SAC, abl-\exda{} degrades the performance of SAC by distilling policy only to augmented observations except the original.   

\begin{center}
    \begin{tabular}{cc}

    \toprule
    Hyperparameter & Value  \\
    \midrule  
        \# of action repeat & 8 \\
        \# of frame stack & 3 \\
        Data augmentation & Random Convolution \\
        
        Learning rate of DA & $10^{-2}$ \\
        Batch size & 128\\
        \# of train steps & 100000 \\
        \# of distill epochs & 10 \\
        Replay buffer capacity & 100000 \\
        Init steps & 1000 \\
        Learning rate of critic & $10^{-3}$ \\
        $\beta$ of critic & 0.9 \\
        $\tau$ of critic & 0.01 \\
        Target update frequency of critic & 2 \\
        Learning rate of actor & $10^{-3}$ \\
        $\beta$ of actor & 0.9 \\
        Log std min of actor & -10 \\
        Log std max of actor & 2 \\
        Update frequency of actor & 2 \\
        Encoder type & pixel \\
        Feature dimension of encoder & 50 \\
        Learning rate of encoder & $10^{-3}$ \\
        $\tau$ of encoder & 0.05 \\
        \# of layers & 4 \\
        \# of filters & 32 \\
        Latent dimension & 128 \\
        Discount factor & 0.99 \\
        Learning rate of $\alpha$ & $10^{-4}$\\
        $\beta$ of $\alpha$ & 0.5 \\
        
    \bottomrule
    \end{tabular}
\end{center}


\end{document}